\documentclass[10pt,twocolumn,letterpaper]{article}

\usepackage{cvpr}              

\usepackage[dvipsnames]{xcolor}

\definecolor{cvprblue}{rgb}{0.21,0.49,0.74}
\usepackage[pagebackref,breaklinks,colorlinks,citecolor=cvprblue]{hyperref}

\usepackage[ruled,linesnumbered]{algorithm2e}
\usepackage{amsthm}
\newtheorem{conclusion}{Conclusion}

\usepackage{multirow}
\usepackage{booktabs}
\usepackage{caption}
\usepackage{setspace}
\def\paperID{4139} 
\def\confName{CVPR}
\def\confYear{2024}

\title{LEAD: Exploring Logit Space Evolution for Model Selection}

\author{{Zixuan Hu{$^{1}$}} \quad Xiaotong Li{$^{1}$} \quad Shixiang Tang{$^{2}$} \quad Jun Liu{$^{3}$} \quad Yichun Hu{$^{1}$} \quad Ling-Yu Duan{$^{1}$}\thanks{Corresponding Author.} \\
\normalsize
	$^{1}$\	Peking University, Beijing, China,  $^{2}$\ \normalsize The Chinese University of Hong Kong, Hongkong, China,\\ \normalsize$^{3}$\ Singapore University of Technology and Design, Singapore \\
	{\tt\small \{hzxuan, hycc, lingyu\}@pku.edu.cn, lixiaotong@stu.pku.edu.cn}, \\
    \tt\small shixiangtang@cuhk.edu.hk, jun\_liu@sutd.edu.sg
}

\begin{document}
\maketitle
\begin{abstract}
The remarkable success of ``pretrain-then-finetune'' paradigm has led to a proliferation of available pre-trained models for vision tasks. This surge presents a significant challenge in efficiently choosing the most suitable pre-trained models for downstream tasks. The critical aspect of this challenge lies in effectively predicting the model transferability by considering the underlying fine-tuning dynamics. Existing methods often model fine-tuning dynamics in feature space with linear transformations, which do not precisely align with the fine-tuning objective and fail to grasp the essential nonlinearity from optimization. To this end, we present LEAD, a finetuning-aligned approach based on the network output of logits. LEAD proposes a theoretical framework to model the optimization process and derives an ordinary differential equation (ODE) to depict the nonlinear evolution toward the final logit state. Additionally, we design a class-aware decomposition method to consider the varying evolution dynamics across classes and further ensure practical applicability. Integrating the closely aligned optimization objective and nonlinear modeling capabilities derived from the differential equation, our method offers a concise solution to effectively bridge the optimization gap in a single step, bypassing the lengthy fine-tuning process. The comprehensive experiments on 24 supervised and self-supervised pre-trained models across 10 downstream datasets demonstrate impressive performances and showcase its broad adaptability even in low-data scenarios.
\end{abstract}    
\vspace{-0.3cm}
\section{Introduction}
\vspace{-0.1cm}
\label{sec:intro}

\begin{figure}[t]
\begin{center}
\includegraphics[width=\linewidth]{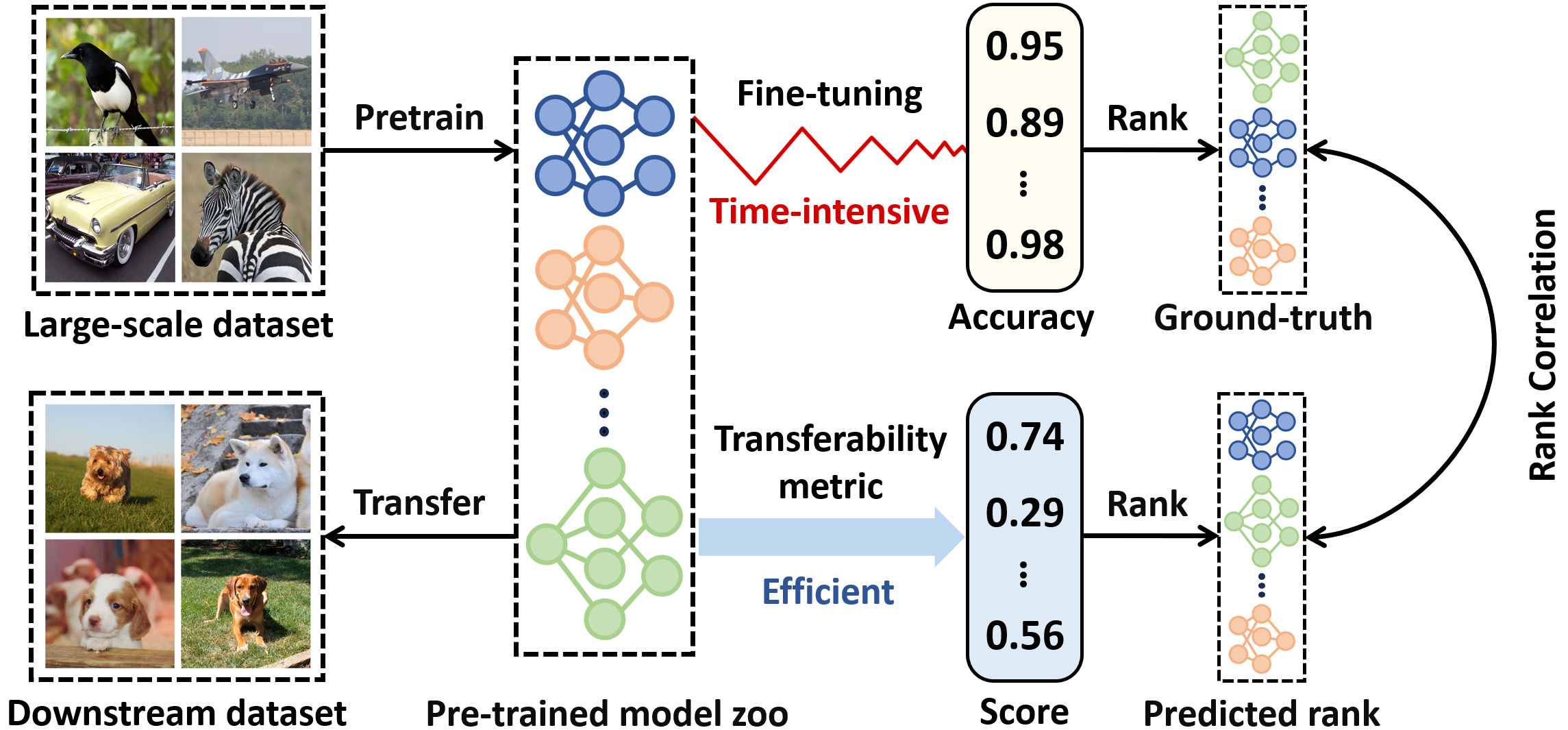}
\end{center}
\caption{\small This diagram illustrates the task of model selection, which aims to rank pre-trained models for choosing the optimal one. The ground-truth ranking sequence is derived from the models' accuracy after fine-tuning. However, due to the time-intensive nature of this process, it requires to design efficient transferability metrics for accurate ranking prediction. The performance of this metric is evaluated using rank correlation, \textit{e.g.,} weighted Kendall's tau \cite{kendall}, to compare the predicted and ground-truth rankings.
}
\label{fig:diagram}
\vspace{-0cm}
\end{figure}

The ``pretrain-then-finetune'' learning paradigm has exhibited remarkable success in a wide range of computer vision tasks~\cite{transfer_better,swav}, leading to an increasing number of diverse pre-trained models open-sourced to the computer vision community. This surge poses a significant challenge in selecting the optimal pre-trained model for a downstream task from a vast model zoo. Considering that the brute-force solution of fine-tuning all models on downstream tasks may be infeasible in many scenarios due to the substantial computational costs, the model selection task emerges \cite{NCE,RSA}, which ranks the transferability of pre-trained models to the given task without incurring much cost, as shown in Fig. ~\ref{fig:diagram}.

Predicting model transferability aims to efficiently estimate the final state after fine-tuning on the downstream task, without network optimization by backward propagation. Prior studies only depend on specific properties of the initial state (\textit{e.g.}, representation similarity \cite{RSA,DDS}, conditional distribution \cite{NCE,nleep}), to serve as transferability indicators. Recent studies reveal that modeling the dynamic process of fine-tuning has become the bottleneck to design more high-quality transferability metrics. Following this philosophy, recent works propose to simulate fine-tuning dynamics on features through multiple stages of linear transformations \cite{ped,sfda}, achieving some promising results.

Despite recent progress in this field, there are still two considerable gaps between existing modeled fine-tuning dynamics and the real ones: (1) Insufficient objective alignment. The designed objectives in prior works do not seamlessly align with fine-tuning, as they often focus more on enhancing class separability via feature manipulation, rather than the classification optimization for both the feature and output spaces. (2) Disregarding nonlinear modeling. Existing methods relying solely on linear transformations cannot well capture the intricate nonlinear changes in the feature space during optimization. Such constraints contribute to growing deviations in dynamic modeling and result in unreliable predictions of the final state.

Delving deeper into the optimization process, we focus on modeling the dynamic changes within the output space (\textit{i.e.,} logits for classification), which not only more closely relates to the fine-tuning objective but also embeds the overall nonlinear dynamics from the network optimization during fine-tuning. Building upon this foundation, we introduce an insightful framework to depict the trajectory of logits toward their final state in a dynamical modeling system. To capture and understand this evolution process of the logit space, the framework provides a discrete iterative optimization equation via gradient analysis and captures nonlinear logit space changes during optimization.

In contrast to the extensive iterations required in fine-tuning optimization, we further extend the framework to propose a new transferability metric ``\textbf{L}ogits \textbf{E}volution \textbf{A}nalysis through \textbf{D}ynamical Equation (LEAD)'', to seize the final evolutionary state in a single step. We perform a mathematical derivation to convert the discrete process into a continuous differential equation and simplify it into an ordinary differential equation (ODE) by leveraging the classical analysis tool, Neural Tangent Kernel (NTK) \cite{NTK}. This ODE concisely models the nonlinear evolution in logit space through its closed-form solution, allowing for real-time estimation of logits. Additionally, we introduce a class-aware decomposition method to enhance prediction accuracy and break the barrier of theoretical assumption of NTK. As a result, our method provides a direct estimation of the evolution by incorporating both the initial model state and the impact of fine-tuning evolution, serving as an effective indicator assessing model transferability.

\par The contributions of this work are three-fold. (1) To the best of our knowledge, we are the first to explore model transferability by modeling the evolution process in the logit space. (2) We propose LEAD, a novel transferability metric with two key components: a theoretical framework providing a closed-form solution to depict the logits evolution towards the final state, and a class-aware decomposition method to guarantee the practical applicability of theoretical findings and improve predictive precision. (3) We validate the effectiveness and generalizability of our method through experiments on 24 different pre-trained models across 10 downstream datasets, achieving SOTA results with an average of $17\%$ gain in the rank correlation. In addition, our method can be naturally extended to low-data scenarios while maintaining competitive performance.

\vspace{-0.1cm}
\section{Related Work}
\vspace{-0.1cm}
\label{sec:relatedworks}
\subsection{Transferability Metric}
\vspace{-0.1cm}
The challenge of efficiently ranking pre-trained models has captured much interest of researchers, leading to the development of various metrics to evaluate the model transferability \cite{leep,NCE,logme,gbc,pac,parc,etran,hscore}. For example, RSA \cite{RSA}, DDS \cite{DDS} and GBC \cite{gbc} are methods based on the distance measurements of the feature space. RSA and DDS assess transferability by evaluating the representation similarity between pre-trained and downstream datasets. GBC employs the Bhattacharyya coefficient to measure class separability in the feature space. $\mathcal{N}$leep \cite{nleep} and Logme \cite{logme} estimate transferability by analyzing the conditional distribution, with $\mathcal{N}$leep utilizing Gaussian mixture model and Logme employing Bayesian optimization. However, these methods treat network outputs as static and overlook the changes introduced by the training process, leading to less effectiveness in scenes with larger fine-tuning dynamics.
\par Recently, SFDA \cite{sfda} and PED \cite{ped} propose to simulate changes in the feature space, respectively projecting the features into a Fisher space and utilizing a physically-inspired model to enhance the class separability, achieving impacts similar to fine-tuning. While these methods achieve some improvements, they solely concentrate on the feature space, disregarding the dynamic changes introduced by the classifier, which is jointly optimized with the backbone. Additionally, they utilize linear transformations to simulate the intricate variations in fine-tuning, resulting in a significant limitation in simulation precision. Instead, our LEAD employs the ordinary differential equation to analyze logit dynamics, maintaining a closely aligned objective and simulating the nonlinear changes in fine-tuning.
\vspace{-0.1cm}
\subsection{Neural Tangent Kernel}
\vspace{-0.1cm}
Studying the behavior of neural network optimization processes has been an enduring challenge due to their nonlinearity and nonconvexity. Neural Tangent Kernel (NTK) \cite{NTK,Finite_WIDTH}, serving as a theoretical tool to characterize the optimization process, has attracted much research attention. Some studies have explored its theoretical properties and applications. Notable works include the NTK's property of remaining constant under infinite network width \cite{NTK}, the explanation of the instability of training \cite{pinn}, the analysis of its bias and variance under finite width \cite{Finite_WIDTH}, and its acceleration algorithm \cite{fastntk}. While NTK can analyze network trainability \cite{NTK,fastntk}, it is constrained by the requirement of infinite width and neglects the consideration of initial logits, resulting in the inability to predict the changes starting from the initial state for a finite-width network. To tackle these issues, we innovatively propose an analysis framework leveraging the property of NTK, suitable for model selection problems to bridge the gap between the initial and final states of finite-width networks.
\begin{figure*}[t]
\begin{center}
\includegraphics[width=\textwidth]{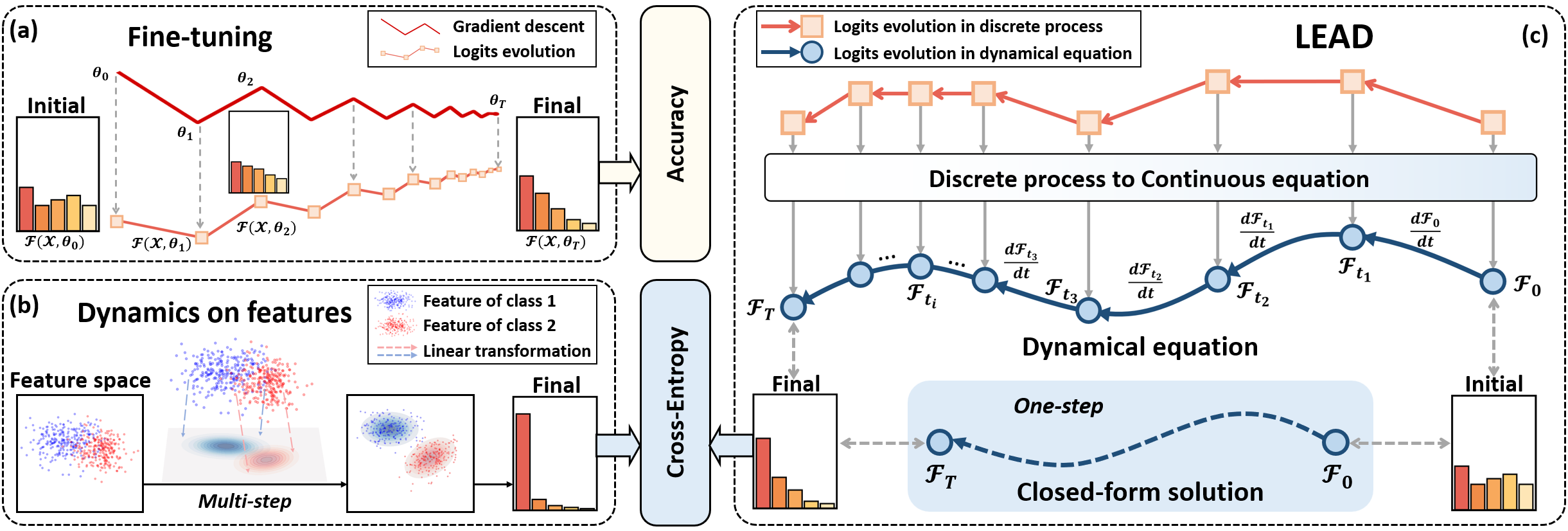}
\caption{\small Comparison of different approaches for modeling the evolution process. (a) Fine-tuning iteratively updates parameters via gradient descent, revealing authentic logits evolution trajectory and classification accuracy for ground-truth ranking. (b) Recent works (\textit{e.g.}, SFDA \cite{sfda}, PED \cite{ped}) perform linear transformations in feature space to enhance class separability, updating logits with modified features. The ranking score is determined by Cross-Entropy loss on updated logits. While these approaches simulate fine-tuning impact on features, they often suffer huge deviations from the authentic results. (c) Our method, LEAD, transforms the discrete logit evolution during fine-tuning into a gradient-based dynamical equation to capture nonlinear changes and can efficiently predict the final state through the closed-form solution. It establishes an analysis framework linking the initial and final states, enabling the precise prediction in one step.}
\label{lead}
\end{center}
\vspace{-0.6cm}
\end{figure*}

\section{Methodology}
In this section, we first present the problem setup for the task of model selection. Then we analyze training mechanisms to model the fine-tuning optimization in a discrete process. Finally, we propose a transferability metric that converts the discrete process into a continuous dynamical equation with a concise solution to directly seize the evolution destination of logits, and feed updated logits into Cross-Entropy loss to assess transferability, as shown in Fig. \protect\ref{lead}.
\vspace{-0.07cm}
\subsection{Problem Setup} 
\vspace{-0.07cm}
Consider $M$ pre-trained models ${\{{\Psi_m}\}_{m=1}^{M}}$ and a downstream dataset $\mathcal{T}=\{x_n, y_n\}_{n=1}^N$ with $K$ classes, where model $\Psi$ consists of its backbone $f$ which outputs the encoded feature, $N$ is the number of images in the downstream task. The purpose of model selection is to design an efficient algorithm to predict the ranking of the performance of $M$ pre-trained models after fine-tuning.

Specifically, given the ground-truth performance rankings $\{G_m\}_{m=1}^M$ of pre-trained models after fine-tuning and the predicted ranking $\{P_m\}_{m=1}^M$ by model selection algorithms, we can leverage the commonly-used rank-based correlation, \emph{i.e.,} the weighted Kendalls' $\tau_w$, to evaluate the effectiveness of the proposed model selection methods, \emph{i.e.,} 
\vspace{-0.15cm}
  \begin{equation}
     \tau_w = \frac{1}{\sum_{i\neq j} w_{ij}}\cdot  \sum_{i\neq j}w_{ij}  \text{sign}(G_i - G_j) \cdot \text{sign}(P_i - P_j) , 
 \end{equation}
where $w_{ij}=\frac{1}{i+j}$ and $\text{sign}(\cdot)$ is the signum function. A larger $\tau_w$ indicates more consistencies between the predicted ranking and the ground truth, which means a better model selection algorithm.

\vspace{-0.07cm}
\subsection{Revisiting the Fine-tuning Optimization} 
\vspace{-0.07cm}
\label{sub:analysis}

When a pre-trained model $\Psi$ transfers to a downstream task $\mathcal{T}:=\{\mathcal{X}, \mathcal{Y}\}$, it undergoes extensive optimization for evolving from its initial state (before fine-tuning) to the final state (after fine-tuning), resulting in complex changes in both feature space and output space. Since the transferability metric is designed to evaluate the final state without fine-tuning, it is essential and challenging to model the optimization process without lengthy computations, bridging the gap between the initial and the final state. To gain insights into the underlying optimization mechanisms, we conduct an analysis to compare the prior arts with fine-tuning to revisit the evolution process toward the final state. 

\par Recent works SFDA \cite{sfda} and PED \cite{ped} (called simulation methods) aim to approximate fine-tuning procedures by simulating feature dynamics. 

They design an objective function $U$ to assess the class separability of the feature space. Then, they apply linear transformations to increase $U$ and feed modified features into a classifier to update logits, as shown below:
\vspace{-0.1cm}
\begin{equation}
\label{eq:simulating}
\begin{split}
    \begin{gathered}
   A, b=\arg\underset{A, b}{\max}\; U(A \cdot f(\mathcal{X})+b, \mathcal{Y}), \\
\Delta \mathcal{F}(\mathcal{X})=\mathcal{C}(A \cdot f(\mathcal{X})+b)-\mathcal{C}(f(\mathcal{X})).
\end{gathered}
\end{split}
\end{equation}
where $A, b$ denote linear transformation parameters, $\mathcal{C}$ denotes a machine learning classifier and $\mathcal{F}$ denotes the function that outputs logits. As a result of the optimization in Eq. (\ref{eq:simulating}), the objective of enhancing class separability is achieved. However, fine-tuning directly minimizes the gap between predicted logits and labels and performs gradient descent on parameters to update the output logits as follows:

\vspace{-0.1cm}
\begin{equation}
\label{eq:finetuning}
\begin{split}
\begin{gathered}
\Delta\theta=-\eta\cdot\frac{\mathrm{d} \mathcal{L}(\mathcal{F}(\mathcal{X}, \theta), \mathcal{Y})}{\mathrm{d} \theta}, \\
\Delta \mathcal{F}  (\mathcal{X})=\mathcal{F}(\mathcal{X}, \theta+\Delta \theta)-\mathcal{F}(\mathcal{X}, \theta) \approx \Delta \theta \cdot \frac{\mathrm{d} \mathcal{F}}{\mathrm{d} \theta} \\
=-\eta\cdot\frac{\mathrm{d} \mathcal{L}}{\mathrm{d} \theta} \cdot \frac{\mathrm{d} \mathcal{F}}{\mathrm{d} \theta}=-\eta\cdot\frac{\mathrm{d} \mathcal{L}}{\mathrm{d} \mathcal{F}}\left(\frac{\mathrm{d} \mathcal{F}}{\mathrm{d} \theta} \cdot \frac{\mathrm{d} \mathcal{F}}{\mathrm{d} \theta}\right).
\end{gathered}
\end{split}
\end{equation}
where $\eta$ denotes the learning rate, $\mathcal{L}$ denotes loss function, and $\theta$ denotes parameters of $\mathcal{F}$. Eq. (\ref{eq:finetuning}) provides the variation of logits within one step, capturing the nonlinear dynamics based on gradients. When comparing the optimization outlined in Eq. (\ref{eq:simulating}) and Eq. (\ref{eq:finetuning}), it becomes evident that simulation methods differ from fine-tuning in terms of objectives and optimization approaches.

\begin{figure}[t]
\begin{center}
\includegraphics[width=1.0\linewidth]{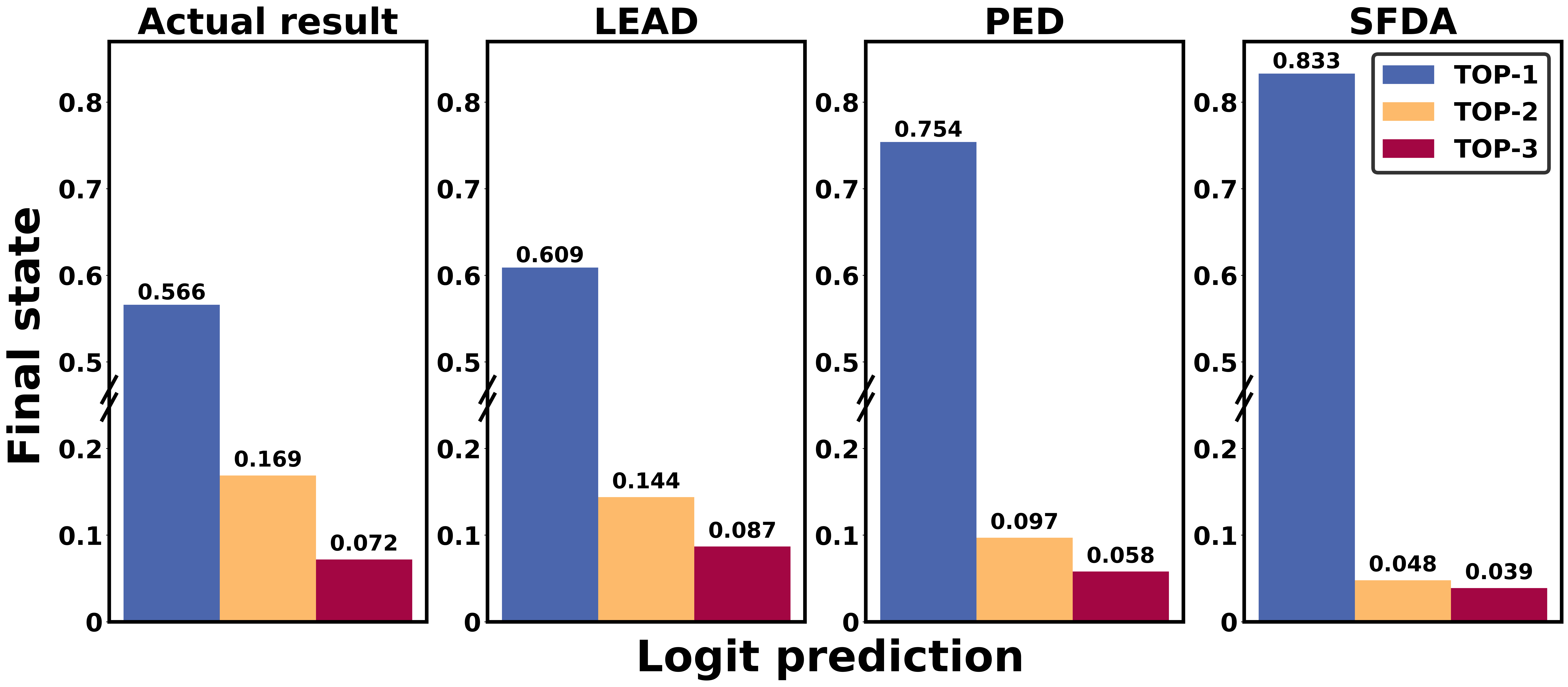}
\end{center}
\setlength{\belowcaptionskip}{-15pt} 
\setlength{\abovecaptionskip}{-2pt} 
\caption{\small Comparison of the average prediction of final state logits across various methods on VOC2007 \cite{voc2007}, where the actual result stems from fine-tuning. Closer to actual results indicates more accurate modeling of the final state. Simulation methods (SFDA, PED) capture the correct trend but exhibit a large deviation from the actual results, affecting the accuracy of model ranking.}
\label{fig:logits}
\end{figure}

Despite preliminary exploration, such simulation methods come with severe limitations due to these inconsistencies, leading to deviations in predicting the final state, as illustrated in Fig. \ref{fig:logits}. Specifically, simulation methods aim to enhance class separability within the feature space, whereas fine-tuning drives logits closer to labels. The insufficient objective alignment introduces distinct inductive bias compared to fine-tuning, resulting in notable variations in modeling dynamics \cite{focal,multi}. Moreover, due to the nonlinear nature of neural networks and complex parameter interactions, linear approximations in simulation methods often fail to accurately capture the dynamic changes \cite{local}. Such limitations hinder the effective modeling of nonlinearity, further constraining modeling precision.

\subsection{Modeling Logits Evolution Process through \quad Dynamical Equation}

To address the aforementioned problems, our study delves deeper into the optimization analysis from Eq. (\ref{eq:finetuning}). In fact, unlike features, the logits derived from the optimization output naturally align with the fine-tuning objectives and reflect the inherent nonlinearity from gradient descent. Therefore, it provides a particularly relevant entry point to analyze dynamical changes during fine-tuning and assess the model transferability through its evolution process.

While the dynamical process of logit space analyzed by Eq. (\ref{eq:finetuning}) remains consistent with fine-tuning, there still lacks an efficient approach for capturing logits evolution, due to the discrete process of iterating over thousands of steps by traditional optimization. To this end, we propose a new transferability metric called ``Logits Evolution Analysis through Dynamical equation (LEAD)'', that bridges the evolution process from the initial logit space towards the final state. In contrast to extensive optimization iterations, our method proposes a dynamical analysis framework for analyzing the nonlinear effects within the logit space, providing an efficient prediction to capture the final state with a concise solution in a single step.

\par\noindent\textbf{Theoretical Background.} 
In our LEAD, we innovatively introduce the Neural Tangent Kernel (NTK) to simplify the dynamical equation. Here, we provide a brief introduction to the definition of the NTK kernel function and the constant-preserving property that we utilize.
\begin{equation}
\begin{aligned}
    \label{eq:infinite}  \hat{\Phi}_{\theta_t}(\mathcal{X}, \mathcal{X}) &\triangleq \nabla_{\theta_t} \mathcal{F}\left(\mathcal{X} ; \theta_t\right) \nabla_{\theta_t} \mathcal{F}\left(\mathcal{X} ; \theta_t\right)^{\top}, \\
        \Phi(&\mathcal{X}, \mathcal{X})\triangleq \lim _{l \rightarrow \infty} \hat{\Phi}_{\theta_t}(\mathcal{X}, \mathcal{X}),
\end{aligned}
\end{equation}
where $\nabla$ denotes the Laplace Operator, and $l$ denotes the root mean square of the widths of all layers in the network. The kernel function $\hat{\Phi}_{\theta_t}$ remains invariant throughout the training process as $l$ approaches infinity with an approximation error of $O(\frac{1}{l^2})$ \cite{NTK,Finite_WIDTH}. For simplicity, in the remainder of this paper, we denote functions at time $t$ as: $\mathcal{F}_t(\cdot)\triangleq\mathcal{F}(\cdot,\theta_t)$, $\hat{\Phi}_t(\cdot)\triangleq\hat{\Phi}_{\theta_t}(\mathcal{X}, \mathcal{X})$, $\Phi \triangleq \Phi(\mathcal{X}, \mathcal{X})$. 
\vspace{-1pt}
\par\noindent\textbf{Dynamical Equation and Closed-form Solution.} 
To bypass the complexity of iterative computations, we extend the analysis in Eq. (\ref{eq:finetuning}) and theoretically derive a dynamical equation that is efficient for analysis. Specifically, we seek for a mathematical derivation through the theory of limit approximation to convert the discrete process into a continuous differential equation, enabling us to track the trajectory of logits in the solution space of the equation. Furthermore, we leverage the property of the NTK in Eq. (\ref{eq:infinite}) to simplify the equation into an Ordinary Differential Equation (ODE) with an initial value condition. Due to space constraints, we directly provide the equation and the detailed proof can be found in the Appendix.
\begin{equation}
\label{eq:ode}
\frac{\mathrm{d}\mathcal{F}_t}{\mathrm{d} t}   =-\eta\cdot \Phi\cdot \frac{\mathrm{d}\mathcal{L}_t}{\mathrm{d}\mathcal{F}_t}, \; \mathcal{F}_0 = log_{init}.
\end{equation}
where $log_{init}$ is the initial state of logits obtained through feeding features and labels of downstream datasets to a classification algorithm. Compared to the discrete process, our proposed ODE depicts the continuous evolution of the fine-tuning dynamics, allowing for a more flexible modeling approach. It transforms discrete gradient effects into equivalent differential effects in continuous time, preserving the capability for modeling nonlinear changes and providing precise prediction of the evolution process. Furthermore, unlike employing numerical methods as the ODEsolver, we can directly obtain its closed-form solution through integration over time dimension, allowing us to bypass the lengthy evolution process and reach the optimization destination in a single step, as shown below:
\begin{equation}
\label{eq:solution}
\mathbb{E}\left(\mathcal{F}_t\left(\mathcal{X}\right)\right)=\left(\mathbb{I}-e^{-\eta \Phi\cdot t}\right) \mathcal{Y}+e^{-\eta \Phi\cdot t} \cdot log_{init}.
\end{equation}
As depicted in Eq. (\ref{eq:solution}), the solution is derived by interpolating between the one-hot label and the initial observation of logits, with the interpolation coefficients determined by the NTK matrix. It comprises two crucial elements: $log_{init}$, representing the value of logits before optimization and determining the start of the evolution process; and $\Phi$, indicating the convergence rate to the label and determining the trajectory of logit evolution. Our proposed framework efficiently models the evolution process, enabling a prediction of real-time changes.

\par\noindent \textbf{Class-aware Decomposition.} 
While Eq. (\ref{eq:solution}) provides the theoretical result for the final state, computing the value of $\Phi$ under finite width remains challenging. Prior studies \cite{structured,Finite_WIDTH} indicate that, even with finite width, the convergence rate can be assessed by using the eigenvalues of the initial state NTK matrix $\hat{\Phi}_0$. Inspired by this empirical finding, we further conduct a decomposition approach to bridge theory and practice, utilizing the eigenvalue decomposition on $\hat{\Phi}_0$ to replace the convergence rate determined by $\Phi$ in Eq. (\ref{eq:solution}) with eigenvalues.

\par Due to the network exhibiting varying classification abilities across different classes, their convergence behavior and evolution dynamics also differ during the fine-tuning process. To enhance the modeling of behaviors in different classes, we separately compute NTK for each class, rather than mixing them together, to decouple the respective optimization effect. Specifically, we extract $n$ samples of each class to calculate their respective NTK matrices and perform decomposition to obtain eigenvalues. The average eigenvalue of each class serves as the unified parameter to determine the interpolation coefficient for this class:
\begin{equation}
\label{eq:class_aware}
\mathbb{E}\left(\mathcal{F}_t\left(x\right)\right)=\left(\mathbb{I}-e^{-\eta \overline{\lambda}\cdot t}\right) y+e^{-\eta \overline{\lambda}\cdot t} \cdot log_{init}.
\end{equation}
where $x, y$ is a given sample and its one-hot label, $\overline{\lambda}$ is the average eigenvalue of the NTK matrix associated with the class of $x$. Building upon the conclusion of Eq. (\ref{eq:solution}), Eq. (\ref{eq:class_aware}) avoids the demand of infinite width, enabling practical computability of the prediction results. As shown in Fig. \ref{lead}, we employ the results of Eq. (\ref{eq:class_aware}) to efficiently obtain predictions for the final state logits. Subsequently, following the common practice \cite{sfda,ped,nleep}, we feed predictions into the Cross-Entropy loss to obtain the transferability score for model ranking.

\par\noindent\textbf{Summary.} Our method explicitly models the fine-tuning optimization through the theoretical analysis framework and provides a concise dynamical solution to bypass the lengthy optimization process in a single step, which seamlessly integrates the initial state observations and the evolution trajectory towards the final state. Therefore, we can efficiently predict the final logit state through this dynamical solution. Through comparing logit prediction accuracy among different models, it can serve as an effective metric for predicting model transferability.

\begin{table*}[h]
\caption{\small Performance comparisons of different transferability metrics on the supervised model zoo. We measure the performance with weighted Kendall’s $\tau_w$ and a larger $\tau_w$ represents a better prediction rank. The best results are in bold and the second-best are in underline.} 
\centering
\label{tab:supervised}
\resizebox{\textwidth}{!}{
\begin{tabular}{ccccccccccccc}
\toprule[1pt]
\multirow{2}{*}{\centering\textbf{Method}} & \multirow{2}{*}{\centering\textbf{Reference}} & \multicolumn{10}{c}{\textbf{Downstream Target Dataset}}  \\
  & & Food  & Caltech  & Flowers  & Cars  & CIFAR100  & DTD  & CIFAR10  & Pets  & SUN397  & VOC2007  & \\
\hline
PARC \cite{parc}     & NeurIPS'21  & 0.363  & 0.422  & 0.066  & 0.135  & -0.092  & 0.536 & 0.310  & 0.114  & -0.097  & 0.707  \\
Logme \cite{logme}    & ICML'21  & 0.336  & 0.349  & 0.425  & 0.538  & 0.603  & \underline{0.651}  & 0.783  & 0.372  & 0.440  & 0.673  \\
$\mathcal{N}$leep \cite{nleep}   & CVPR'21  & 0.504  & \underline{0.678}  & 0.292  & 0.435  & 0.686  & 0.452  & 0.656  & \underline{0.686}  & \textbf{0.770}  & 0.645  \\
PACTran \cite{pac} & ECCV'22  & 0.709  & 0.372  & 0.480  & 0.151  & 0.742  & 0.372  & 0.757  & 0.451  & 0.422  & 0.536   \\
SFDA  \cite{sfda}   & ECCV'22  & 0.534  & 0.597  & 0.421  & 0.472  & 0.749  & 0.421  & \underline{0.817}  & 0.625  & 0.562  & \underline{0.728}   \\
GBC  \cite{gbc}   & CVPR'22  & \underline{0.844}  & 0.430  & 0.348  & \textbf{0.735}  & \underline{0.830}  & 0.480  & \textbf{0.866}  & 0.358  & 0.574  & 0.654   \\
ETran  \cite{etran}  & ICCV'23  & 0.709  & 0.627  & \underline{0.534}  & 0.557  & 0.749  & 0.605  & 0.719  & 0.651  & 0.577  & 0.673   \\
PED  \cite{ped}    & ICCV'23  & 0.573  & 0.597  & 0.451  & 0.472  & 0.749  & 0.421  & 0.770  & 0.625  & 0.562  & \underline{0.728}   \\
\hline
LEAD    & This paper  & \textbf{0.892}  & \textbf{0.698}  & \textbf{0.786}  & \underline{0.586}  & \textbf{0.835}  & \textbf{0.689}  & 0.791  & \textbf{0.841}  & \underline{0.609}  & \textbf{0.743}  \\
\bottomrule[1pt]
\end{tabular}
}
\vspace{-0.3cm}
\end{table*}
\vspace{-0.1cm}
\section{Experiments}
\vspace{-0.1cm}
To demonstrate the effectiveness and robustness of our approach, we evaluate LEAD on diverse pre-trained models, including both supervised and self-supervised convolutional neural networks. Our evaluation spans 10 widely used classification benchmarks in transfer learning, employing weighted Kendall's $\tau_w$ as rank correlation to evaluate the transferability metric. Additionally, we highlight the effectiveness of LEAD in the low-data regime, where only $2/5/10$ samples are available per class.
\vspace{-0.05cm}
\subsection{Implementation Details}
\vspace{-0.05cm}
Given a pre-trained model $\Psi$, we append a random-initialized \cite{kaiming} MLP $h$ as a classification head after its backbone $f$, and combine them as $\mathcal{F}:=h\circ f$ to produce logits. To compute the logits in the final state, we only need to derive the NTK matrix of each class and $log_{init}$ through Eq. (\ref{eq:class_aware}). We follow the approach in \cite{fastntk} to compute the NTK matrix and employ a robust and efficient machine learning classifier, the multi-class SVM \cite{SVM}, to generate $log_{init}$. Due to space constraints, we provide a more detailed explanation and pseudocode in the Appendix. 
\vspace{-0.05cm}
\subsection{Benchmarks}
\vspace{-0.05cm}
\textbf{Target Datasets.} We employ 10 classification benchmarks commonly utilized in transfer learning research including Caltech-101 \cite{caltech101}, Stanford Cars \cite{cars}, CIFAR10 \cite{cifar}, CIFAR100 \cite{cifar}, DTD \cite{dtd}, Oxford 102 Flowers \cite{flowers},  Food-101 \cite{food}, Oxford-IIIT Pets \cite{pets}, SUN397 \cite{sun397}, and VOC2007 \cite{voc2007}. These datasets encompass abundant characteristics, including backgrounds, textures, and coarse/fine-grained scenes, spanning diverse fields.
\par\noindent\textbf{Ground Truth Rank.} To conduct the ground truth ranking for the model zoo, we follow the approach in \cite{sfda} and \cite{logme}. Specifically, we employ a grid search strategy to determine the actual performance of each model in the downstream task. This strategy involves trying various learning rates from the set  $\{10^{-1},10^{-2},10^{-3},10^{-4}\}$ and different weight decay values from the set $\{0,10^{-6},10^{-5},10^{-4},10^{-3}\}$. To ensure reliability, we repeat each experiment 5 times with different random seeds and take the average as the result.

\vspace{-0.05cm}
\subsection{Evaluation on Supervised Models}
\vspace{-0.05cm}
\textbf{Model Zoo.} We construct a Model Zoo that includes 12 CNN models supervised pre-trained on ImageNet \cite{imagenet}, covering widely-used architectures: ResNet-34 \cite{resnet}, ResNet-50 \cite{resnet}, ResNet-101 \cite{resnet}, ResNet-152 \cite{resnet}, DenseNet-121 \cite{densenet}, DenseNet-161 \cite{densenet}, DenseNet-169 \cite{densenet}, DenseNet-201 \cite{densenet}, MNet-A1 \cite{mnasnet}, MobileNetV2 \cite{mobilenetv2}, GoogleNet \cite{googlenet}, and InceptionV3 \cite{inceptionv3}. The accuracy results of these models after fine-tuning can be found in the Appendix.
\par\noindent\textbf{Result Analysis.} To show the effectiveness of LEAD, we compare it with prior arts, such as Bayesian-based Logme, Separability-based GBC, and simulation methods SFDA and PED. As shown in Tab. \ref{tab:supervised}, simulation methods consider underlying dynamic effects and exhibit promising performances on some datasets. However, as the aforementioned analysis in Sec. \ref{sub:analysis} shows, the bias in the simulation process causes these methods to have unstable improvements, particularly resulting in unsatisfactory performance on Flowers, Food, and Pets. Benefiting from a precise analysis of the evolution, our LEAD yields a performance gain of +0.335, +0.319, +0.216 compared to PED on these three datasets, respectively. Overall, LEAD achieves the best $\tau_w$ on 7 target datasets and obtains an average $\tau_w$ of 0.747 which is relatively $18\%$ better than the latest SOTA ETran \cite{etran}. Our experiments highlight the importance of accurate modeling of dynamics and confirm the effectiveness of our method to model the evolution process through the dynamical equation.
\vspace{-0.05cm}
\subsection{Evaluation on Self-supervised Models}
\vspace{-0.05cm}
\label{sec:self-supervised model zoo}
\textbf{Model Zoo.} Self-supervised learning models (SSL models) have shown their efficacy in transfer learning and often display superior generalization performance compared to Supervised models. To illustrate the broad applicability of our approach, inspired by \cite{sfda,ped}, we assemble a self-supervised model zoo trained on ImageNet with BYOL \cite{byol}, Infomin \cite{infomin}, PCLv1 \cite{pcl}, PCLv2 \cite{pcl}, Selav2 \cite{sela}, InsDis \cite{indis}, SimCLRv1 \cite{simclr-v1}, SimCLRv2 \cite{simclr-v2}, MoCov1 \cite{moco-v1}, MoCov2 \cite{moco-v2}, DeepClusterv2 \cite{deepcluster}, and SWAV \cite{swav}.
\par\noindent\textbf{Result Analysis.} Due to differing pre-training and downstream classification objectives, SSL models undergo larger dynamic changes during the fine-tuning process, leading to distinct transferability properties compared to supervised models \cite{ssl_transfer}. As shown in Tab. \ref{tab:supervised} and \ref{tab:self-supervised}, methods that disregard dynamics (\textit{e.g.} PARC, $\mathcal{N}$leep) perform competitively in the supervised model zoo, but fail on some datasets for SSL models. For example, $\mathcal{N}$leep and PARC have a negative small rank correlation on VOC2007 (-0.101) and CIFAR100 (-0.136), respectively. To maintain fairness, we refrain from comparing with GBC \cite{gbc} in the SSL model zoo, which is designed for supervised scenarios. On the contrary, SFDA and PED exhibit better performances due to consideration of training dynamics. Nevertheless, the limitation of the simulating process still hinders their performance. Compared to them, LEAD achieves a consistent improvement on 7 datasets and obtains an average $\tau_w$ of 0.746 which is relatively $15\%$ better than PED (0.648). The results validate the effectiveness of LEAD for SSL models.

\begin{table*}[h]
\caption{Performance comparisons on the self-supervised model zoo. The best results are in bold and the second-best are in underline.}
\centering
\label{tab:self-supervised}
\resizebox{\textwidth}{!}{
\begin{tabular}{ccccccccccccc}
\toprule[1pt]
\multirow{2}{*}{\centering\textbf{Method}} &\multirow{2}{*}{\centering\textbf{Reference}} & \multicolumn{10}{c}{\textbf{Downstream Target Dataset}}  \\
&   & Food  & Caltech  & Flowers  & Cars  & CIFAR100  & DTD  & CIFAR10  & Pets  & SUN397  & VOC2007  & \\
\hline
PARC \cite{parc}    & NeurIPS'21    & 0.359         & 0.196         & 0.622          & 0.424          & -0.136         & 0.447          & 0.147         & 0.496          & -0.006        & 0.618         \\
Logme \cite{logme}  & ICML'21    & 0.570          & 0.051         & 0.604          & 0.375          & -0.008         & 0.627          & 0.295         & \underline{0.684}          & 0.217         & 0.158          \\
$\mathcal{N}$leep \cite{nleep}  & CVPR'21    & 0.574         & 0.525         & 0.534          & 0.486          & 0.276          & 0.641          & -0.044        & \textbf{0.792} & \underline{0.719}         & -0.101          \\
PACTran \cite{pac} & ECCV'22    & \underline{0.720}          & \underline{0.622}         & 0.601          & 0.474          & 0.529          & 0.614          & 0.477         & 0.641          & 0.638         & \underline{0.620}                 \\
SFDA  \cite{sfda}  & ECCV'22    & 0.685         & 0.523         & \underline{0.749}          & 0.515          & 0.548          & 0.773          & 0.619         & 0.586          & 0.698         & 0.568                 \\
ETran \cite{etran}  & ICCV'23    & 0.465         & 0.405         & 0.644          & 0.587          & \underline{0.650}           & 0.214          & 0.606         & 0.338          & 0.520          & 0.376                  \\
PED  \cite{ped}  & ICCV'23    & 0.581         & 0.614         & \textbf{0.777} & \underline{0.649}          & 0.568          & \textbf{0.907} & \underline{0.673}         & 0.462          & 0.661         & 0.583                 \\
\hline
LEAD   & This paper & \textbf{0.860} & \textbf{0.780} & 0.725          & \textbf{0.663} & \textbf{0.776} & \underline{0.825}          & \textbf{0.713} & 0.629          & \textbf{0.760} & \textbf{0.723}  \\
\bottomrule[1pt]
\end{tabular}
}
\vspace{-0.15cm}
\end{table*}

\begin{table*}[h]
\caption{\small Performance comparisons with different $N/K$, where $K$ is the number of classes, $N$ is the number of samples for computing the metric. We divide 10 benchmarks into two groups according to the number of classes: 100+ classes and 10-99 classes, and evaluate the average of $\tau_w$ on the two groups. Additionally, we also provide the average results on the 10 benchmarks, recorded as Avg. The best results are denoted in bold, the second best results are denoted in underline, and - denotes cannot give a valid ranking.}
\centering
\label{tab:lowdata}
\resizebox{\textwidth}{!}{
\begin{tabular}{cccccccccc}
\toprule[1pt]
\multicolumn{10}{c}{Supervised / Self-supervised pre-trained Model} \\ \hline
\multicolumn{1}{c|}{\multirow{2}{*}{Method}} & \multicolumn{3}{c|}{$N/K=2$} & \multicolumn{3}{c|}{$N/K=5$} & \multicolumn{3}{c}{$N/K=10$} \\ \cline{2-10} 
\multicolumn{1}{c|}{} & 100+ classes & 10-99 classes & \multicolumn{1}{c|}{Avg} & 100+ classes & 10-99 classes & \multicolumn{1}{c|}{Avg} & 100+ classes & 10-99 classes & Avg \\ \hline
\multicolumn{1}{c|}{GBC} & 0.17 / 0.32 & 0.10 / 0.17 &  \multicolumn{1}{c|}{0.13 / 0.25} & 0.45 / 0.40 & 0.22 / 0.18 &  \multicolumn{1}{c|}{0.36 / 0.31} & \underline{0.59} / 0.42 & 0.42 / 0.22 & 0.52 / 0.34 \\
\multicolumn{1}{c|}{PACTran} & \underline{0.43} / \textbf{0.49} & \underline{0.20} / 0.39 &  \multicolumn{1}{c|}{\underline{0.33} / \underline{0.45}} & 0.42 / \underline{0.48} & 0.31 / \underline{0.44} &  \multicolumn{1}{c|}{0.38 / \underline{0.47}} & 0.56 / \underline{0.52} & 0.48 / 0.46 & 0.53 / \underline{0.50} \\
\multicolumn{1}{c|}{ETran} & 0.42 / 0.43 & 0.12 / \underline{0.41} &  \multicolumn{1}{c|}{0.29 / 0.42} & \textbf{0.48} / 0.47 & \underline{0.31} / 0.38 &  \multicolumn{1}{c|}{\underline{0.41} / 0.43} & 0.56 / 0.47 & \underline{0.56} / 0.48 & \underline{0.56} / 0.48 \\
\multicolumn{1}{c|}{PED} & - / 0.36 & - / 0.24 &  \multicolumn{1}{c|}{- / 0.31} & - / 0.43 & - / 0.35 &  \multicolumn{1}{c|}{- / 0.40} & 0.30 / 0.50 & 0.22 / \underline{0.56} & 0.27 / 0.52 \\ \hline
\multicolumn{1}{c|}{LEAD} & \textbf{0.45} / \underline{0.48} & \textbf{0.51} / \textbf{0.45} &  \multicolumn{1}{c|}{\textbf{0.47} / \textbf{0.46}} & \textbf{0.48} / \textbf{0.49} & \textbf{0.64} / \textbf{0.53} &  \multicolumn{1}{c|}{\textbf{0.54} / \textbf{0.51}} & \textbf{0.65} / \textbf{0.53} & \textbf{0.64} / \textbf{0.60} & \textbf{0.64} / \textbf{0.56} \\ 
\bottomrule[1pt]
\end{tabular}
}
\vspace{-0.3cm}
\end{table*}
\vspace{-0.15cm}
\subsection{Evaluation on Low-data Regime}
\vspace{-0.1cm}
\textbf{Experimental Settings.} In practical application, privacy and resource constraints may limit our access to the entire downstream dataset. To explore the utilization boundary, we evaluate whether LEAD can provide effective model ranking only using limited samples. We follow three data settings in PACTran \cite{pac} with increasing average number of samples per class $\{2, 5, 10\}$ and divide 10 datasets into two groups according to the number of classes: 100+ classes, including Food, Caltech, Flowers, Cars, CIFAR100, and SUN397; 10-99 classes, including DTD, CIFAR10, Pets, and VOC2007. For each setting, we repeatedly sample 5 times for evaluation and take the average as the result.

\par\noindent \textbf{Result Analysis.}
As shown in Tab. \ref{tab:lowdata}, all methods, including PACTran designed for this scenario, exhibit significant performance declines under low-data settings. For instance, ETran's performance in the supervised model zoo drops from 0.64 with the entire dataset to 0.29, 0.41, and 0.56 in three low-data settings. Moreover, some methods, such as SFDA and PED fail to provide prediction rankings in some settings. For example, when $N/K=2, 5$, PED cannot correctly evaluate class separability due to insufficient data, imposing huge changes to completely separate different classes. This leads to all models perfectly classifying all samples, resulting in identical scores that cannot be ranked. In contrast, our LEAD achieves stable results, ensuring $\tau_w \geq 0.45$ even in extreme settings like $N/K=2$, and consistently outperforms other methods. For example, under $N/K=10$, its average $\tau_w$ in the supervised model zoo reaches 0.64, which is relatively $14\%$ better than the latest SOTA ETran (0.56) and is equivalent to the latest SOTA on the complete dataset, with less than $5\%$ of data size. The results demonstrate the robustness of our method in the face of data limitations.
\vspace{-0.3cm}
\section{Ablation Study and Efficiency Analysis}
\vspace{-0.15cm}
In this section, without loss of generality, we conduct ablation studies on Caltech, CIFAR10, and VOC2007, building upon the evaluation of SSL models in Sec. \ref{sec:self-supervised model zoo} for saving time. We conduct separate experiments to analyze the impact of interpolation coefficients, a key component of LEAD, aiming to delve into the influence of hyperparameters and implementation details, gaining insights into key factors contributing to LEAD's effectiveness. Additionally, we provide a runtime comparison to verify efficiency, and more ablation study results are provided in the appendix.

\vspace{-0.1cm}
\subsection{Interpolation Coefficient}
\vspace{-0.1cm}
The interpolation coefficients are determined by time coefficient $t$ and the NTK matrix. We conduct ablation experiments on these two key factors:

\par
\begin{figure}[h]
\begin{center}
\includegraphics[width=0.93\linewidth]{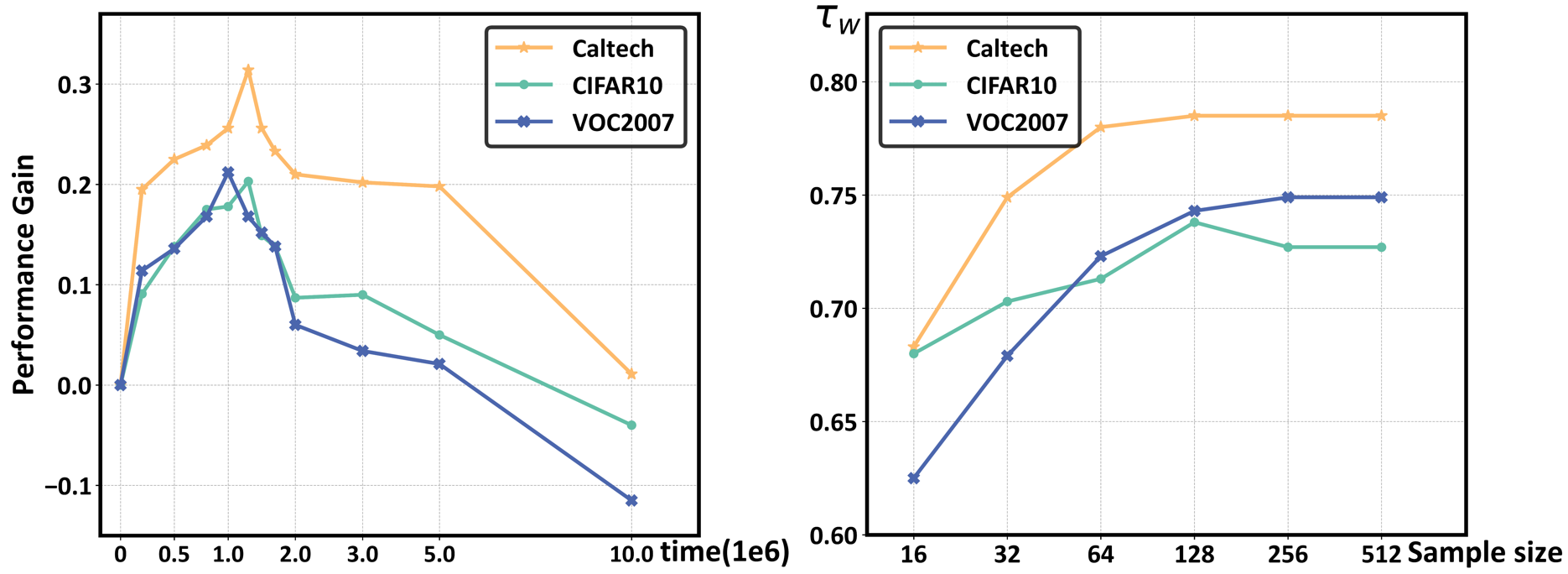}
\setlength{\belowcaptionskip}{-0.4cm} 
\setlength{\abovecaptionskip}{0pt} 
\caption{\small The left picture shows the performance gain in $\tau_w$ with respect to different time $t$ and the right shows the performance in different sample sizes for each class NTK.}
\label{fig:time_ablation}
\end{center}
\vspace{-0.29cm}
\end{figure}

\noindent\textbf{Time Coefficient $t$.}
As we model the discrete process as a continuous ODE, the continuous variable $t$ is proportional to the number of updates, determining the progress of the evolution process. In Fig. \ref{fig:time_ablation}, we evaluate our method under varied time settings, keeping the unit as $1 \mathrm{e} 6$ consistent across all experiments. It is evident that performance improves as the time increases within a specific range. However, if the dynamic effect persists for an extended period (e.g., 10.0), all model predictions converge too closely to the ground-truth label, reducing discrimination for model ranking. Hence, we set the time coefficient as 1 by default.

\par\noindent\textbf{Sample Size for NTK.}
To enhance the perception of different dynamic intensities of classes, we propose the class-aware decomposition that employs $S$ samples of each class to obtain NTK matrix and perform spectrum analysis to derive eigenvalues. In Fig. \ref{fig:time_ablation}, we conduct ablation experiments to explore the effect of varying $S$. We observe that performance generally improves as $S$ increases, and gradually reaches stability. This demonstrates that when the sample size reaches a certain size, eigenvalues reliably capture differences in dynamic intensity, offering a refined analysis result of the dynamics. Therefore, we set $S=64$ to ensure the method's efficiency and stability in all experiments.

\vspace{-0.1cm}
\subsection{Running Time Comparison}
\vspace{-0.1cm}
In experiment results, we have demonstrated the effectiveness of LEAD in various scenarios. In this section, we highlight the computational efficiency of LEAD. With the concise solution in Eq. (\ref{eq:class_aware}), LEAD predicts the final state only requiring the NTK matrix and $log_{init}$. As shown in Tab. \ref{table:running time}, the running time of LEAD is about half that of simulation methods (45s compared to 92s), ranking second after GBC which displays unstable performance. Notably, LEAD takes 3s to compute the NTK, while feature extraction takes 420s, resulting in only a marginal $0.7\%$ increase in GPU time. The time to compute $log_{init}$ is 42s CPU time, indicating significant potential for exploring faster classification algorithms to further reduce the overall time.
\vspace{-0.2cm}

\begin{table}[h]
\begin{center}
\setlength{\belowcaptionskip}{-3pt} 
\setlength{\abovecaptionskip}{-3pt} 
\caption{\small The comparisons of average running time on CIFAR10.}
\label{table:running time}
\resizebox{\linewidth}{!}{
\begin{tabular}{c|cccccc|c}
\toprule[1pt]
Metrics & PARC &  $\mathcal{N}$LEEP & PACTran & SFDA  & GBC & PED  & Ours \\ \hline
Running Time  & 111s  & 1430s & 77s  & 92s  &  26s   & 97s & 45s\\
\toprule[1pt]
\end{tabular}}
\end{center}
\vspace{-0.7cm}
\end{table}
\vspace{-0.40cm}
\section{Visualization and Analysis}
\vspace{-0.15cm}
In this section, we validate our approach through some visualizations and these results clearly illustrate how our model effectively addresses the model selection problem.
\vspace{-0.07cm}
\subsection{Logit Prediction}
\begin{figure}[h]
\begin{center}
\includegraphics[width=0.92\linewidth]{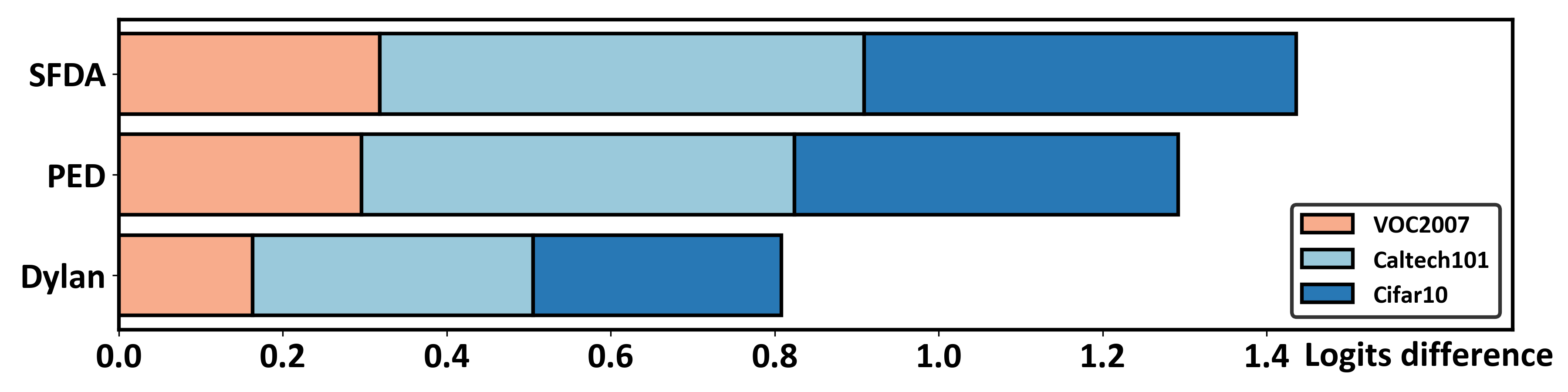}
\end{center}
\setlength{\belowcaptionskip}{-15pt} 
\setlength{\abovecaptionskip}{-10pt} 
\caption{\small Comparison of the average difference between logits predicted by different methods and logits after fine-tuning. The difference is calculated utilizing the Euclidean distance.
}
\label{fig:final}
\vspace{-0.1cm}
\end{figure}

\vspace{-0.07cm}
We have conducted a comparison of the predicted logits distribution on VOC2007 using different methods, as illustrated in Fig. \ref{fig:logits}. Notably, predictions obtained by LEAD closely match the authentic distribution after fine-tuning. To provide further quantitative analysis, we visualize the average bias in predicted logits in Fig. \ref{fig:final}. Across three datasets, our method exhibits a substantial reduction compared to simulation methods, achieving nearly a $50\%$ overall reduction. This visualization highlights the precision of our dynamical equation-based approach in capturing logit evolution for more accurate ranking sequences.
\vspace{-0.17cm}

\vspace{-0.05cm}
\subsection{Rank Variation}
\begin{figure}[h]
\begin{center}
\includegraphics[width=0.99\linewidth]{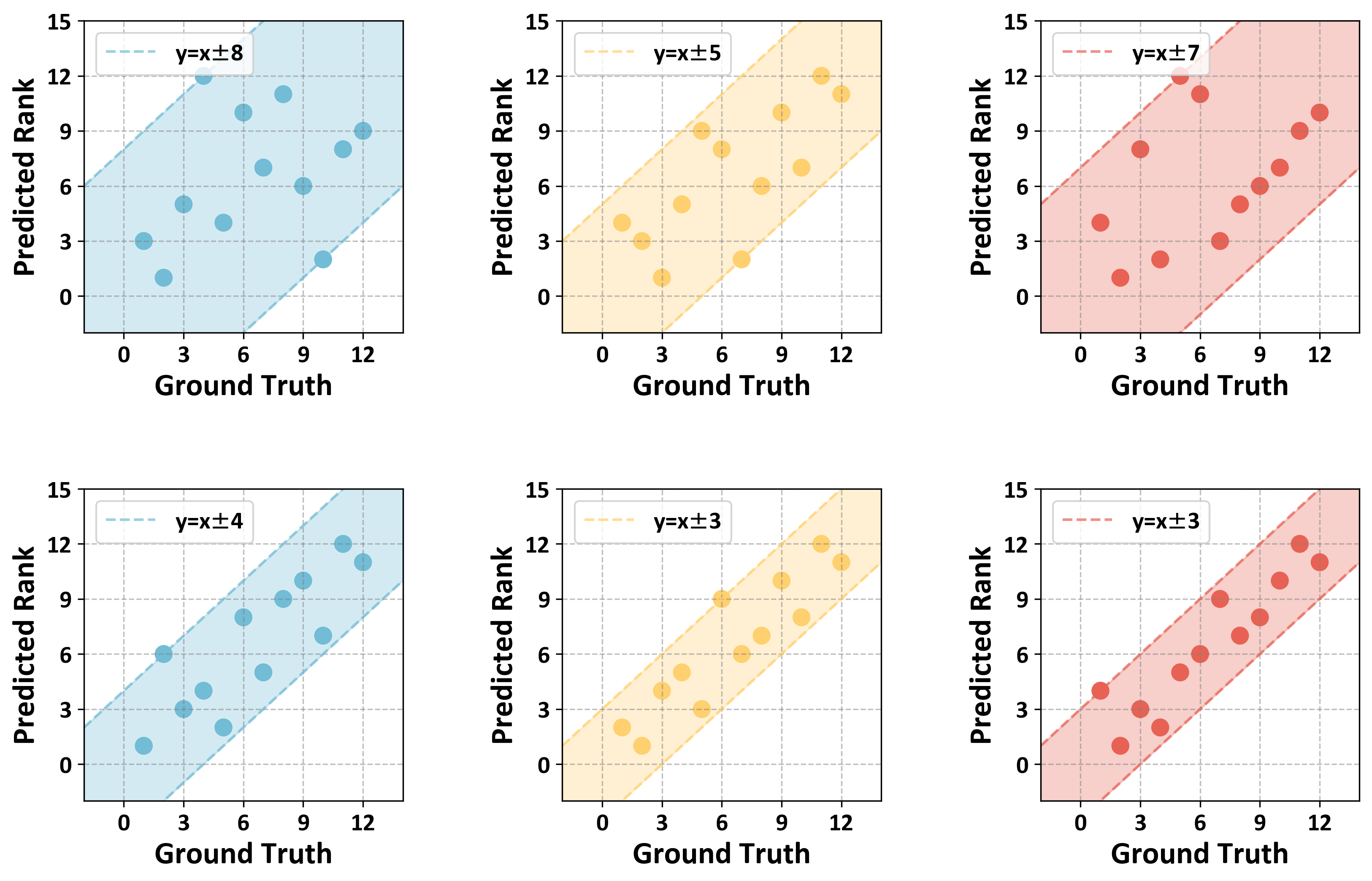}
\setlength{\belowcaptionskip}{-0.5cm} 
\caption{\small The variation in model ranking, utilizing initial states (first row) and predicted final states of LEAD (second row). Results span three datasets, from left to right: CIFAR10, Caltech, and VOC2007. Shaded areas reflect the range of the ranking error.}
\setlength{\belowcaptionskip}{-10pt} 
\setlength{\abovecaptionskip}{-10pt} 
\label{fig:scatter}
\end{center}
\vspace{-0.2cm}
\end{figure}

\vspace{-0.15cm}
By employing our method, we can refine models that have poor initial observation but show better progress during training. To assess the effectiveness of our approach, we conduct a visualization comparing the model rankings between the initial observations and our predicted final state, as shown in Fig. \ref{fig:scatter}. The results clearly illustrate a substantial enhancement in the calibration of model rankings when using our refined observations, with the optimal model now positioned closely to the reference line in the figure. Notably, models starting from an initially unfavorable point can be rapidly elevated to achieve a superior ranking, highlighting the effectiveness of our methodology.
\vspace{-0.15cm}

\vspace{-0.4cm}
\section{Conclusions and Future Work}
\vspace{-0.2cm}
In this work, we propose a dynamical equation designed to capture the evolution process within the logit space, offering an efficient solution for assessing model transferability. We construct a theoretical analysis framework and further devise a more practical and precise solution. And we demonstrate its consistent effectiveness across various scenarios including supervised and self-supervised, full-data and low-data settings. In future work, we intend to expand the current analysis framework beyond classification to cover a broader range of downstream tasks. We hope that our work can enhance the understanding of optimization mechanisms in deep neural networks and shed light on other fields, such as the design of optimization algorithms.
\vspace{-0.1cm}
\vspace{-0.4cm}
\section*{Acknowledgement}
\vspace{-0.2cm}
\label{acknowlegement}
This work was supported by the National Natural Science Foundation of China under Grant 62088102, in part by the PKU-NTU Joint Research Institute (JRI) sponsored by a donation from the Ng Teng Fong Charitable Foundation and in part by AI Joint Lab of Future Urban Infrastructure sponsored by Fuzhou Chengtou New Infrastructure Group and Boyun Vision Co. Ltd. 

{
\small
\bibliographystyle{ieeenat_fullname}
\bibliography{main}

\begin{thebibliography}{55}
\providecommand{\natexlab}[1]{#1}
\providecommand{\url}[1]{\texttt{#1}}
\expandafter\ifx\csname urlstyle\endcsname\relax
  \providecommand{\doi}[1]{doi: #1}\else
  \providecommand{\doi}{doi: \begingroup \urlstyle{rm}\Url}\fi

\bibitem[Asano et~al.(2019)Asano, Rupprecht, and Vedaldi]{sela}
Yuki~Markus Asano, Christian Rupprecht, and Andrea Vedaldi.
\newblock Self-labelling via simultaneous clustering and representation learning.
\newblock \emph{arXiv}, 2019.

\bibitem[Bao et~al.(2019)Bao, Li, Huang, Zhang, Zheng, Zamir, and Guibas]{hscore}
Yajie Bao, Yang Li, Shao-Lun Huang, Lin Zhang, Lizhong Zheng, Amir Zamir, and Leonidas Guibas.
\newblock An information-theoretic approach to transferability in task transfer learning.
\newblock In \emph{ICIP}, pages 2309--2313, 2019.

\bibitem[Bolya et~al.(2021)Bolya, Mittapalli, and Hoffman]{parc}
Daniel Bolya, Rohit Mittapalli, and Judy Hoffman.
\newblock Scalable diverse model selection for accessible transfer learning.
\newblock In \emph{NeurIPS}, pages 19301--19312, 2021.

\bibitem[Bossard et~al.(2014)Bossard, Guillaumin, and Van~Gool]{food}
Lukas Bossard, Matthieu Guillaumin, and Luc Van~Gool.
\newblock Food-101--mining discriminative components with random forests.
\newblock In \emph{ECCV}, 2014.

\bibitem[Caron et~al.(2018)Caron, Bojanowski, Joulin, and Douze]{deepcluster}
Mathilde Caron, Piotr Bojanowski, Armand Joulin, and Matthijs Douze.
\newblock Deep clustering for unsupervised learning of visual features.
\newblock In \emph{ECCV}, pages 132--149, 2018.

\bibitem[Caron et~al.(2020)Caron, Misra, Mairal, Goyal, Bojanowski, and Joulin]{swav}
Mathilde Caron, Ishan Misra, Julien Mairal, Priya Goyal, Piotr Bojanowski, and Armand Joulin.
\newblock Unsupervised learning of visual features by contrasting cluster assignments.
\newblock In \emph{NeurIPS}, pages 9912--9924, 2020.

\bibitem[Chen et~al.(2020{\natexlab{a}})Chen, Kornblith, Norouzi, and Hinton]{simclr-v1}
Ting Chen, Simon Kornblith, Mohammad Norouzi, and Geoffrey Hinton.
\newblock A simple framework for contrastive learning of visual representations.
\newblock In \emph{ICML}, 2020{\natexlab{a}}.

\bibitem[Chen et~al.(2020{\natexlab{b}})Chen, Kornblith, Swersky, Norouzi, and Hinton]{simclr-v2}
Ting Chen, Simon Kornblith, Kevin Swersky, Mohammad Norouzi, and Geoffrey~E Hinton.
\newblock Big self-supervised models are strong semi-supervised learners.
\newblock In \emph{NeurIPS}, pages 22243--22255, 2020{\natexlab{b}}.

\bibitem[Chen et~al.(2020{\natexlab{c}})Chen, Fan, Girshick, and He]{moco-v2}
Xinlei Chen, Haoqi Fan, Ross Girshick, and Kaiming He.
\newblock Improved baselines with momentum contrastive learning.
\newblock \emph{arXiv}, 2020{\natexlab{c}}.

\bibitem[Cimpoi et~al.(2014)Cimpoi, Maji, Kokkinos, Mohamed, and Vedaldi]{dtd}
Mircea Cimpoi, Subhransu Maji, Iasonas Kokkinos, Sammy Mohamed, and Andrea Vedaldi.
\newblock Describing textures in the wild.
\newblock In \emph{CVPR}, pages 3606--3613, 2014.

\bibitem[Deng et~al.(2009)Deng, Dong, Socher, Li, Li, and Fei-Fei]{imagenet}
Jia Deng, Wei Dong, Richard Socher, Li-Jia Li, Kai Li, and Li Fei-Fei.
\newblock Imagenet: A large-scale hierarchical image database.
\newblock In \emph{CVPR}, pages 248--255, 2009.

\bibitem[Ding et~al.(2022)Ding, Chen, Levinboim, Changpinyo, and Soricut]{pac}
Nan Ding, Xi Chen, Tomer Levinboim, Soravit Changpinyo, and Radu Soricut.
\newblock Pactran: Pac-bayesian metrics for estimating the transferability of pretrained models to classification tasks.
\newblock In \emph{ECCV}, pages 252--268. Springer, 2022.

\bibitem[Dwivedi and Roig(2019)]{RSA}
Kshitij Dwivedi and Gemma Roig.
\newblock Representation similarity analysis for efficient task taxonomy transfer learning.
\newblock In \emph{CVPR}, pages 12387--12396, 2019.

\bibitem[Dwivedi et~al.(2020)Dwivedi, Huang, Cichy, and Roig]{DDS}
Kshitij Dwivedi, Jiahui Huang, Radoslaw~Martin Cichy, and Gemma Roig.
\newblock Duality diagram similarity: a generic framework for initialization selection in task transfer learning.
\newblock In \emph{ECCV}, pages 497--513, 2020.

\bibitem[Ericsson et~al.(2021)Ericsson, Gouk, and Hospedales]{ssl_transfer}
Linus Ericsson, Henry Gouk, and Timothy~M Hospedales.
\newblock How well do self-supervised models transfer?
\newblock In \emph{CVPR}, pages 5414--5423, 2021.

\bibitem[Everingham et~al.(2010)Everingham, Van~Gool, Williams, Winn, and Zisserman]{voc2007}
Mark Everingham, Luc Van~Gool, Christopher~KI Williams, John Winn, and Andrew Zisserman.
\newblock The pascal visual object classes (voc) challenge.
\newblock \emph{IJCV}, 88:\penalty0 303--338, 2010.

\bibitem[Fei-Fei(2004)]{caltech101}
Li Fei-Fei.
\newblock Learning generative visual models from few training examples.
\newblock In \emph{CVPR workshop}, 2004.

\bibitem[Gholami et~al.(2023)Gholami, Akbari, Wang, Kamranian, and Zhang]{etran}
Mohsen Gholami, Mohammad Akbari, Xinglu Wang, Behnam Kamranian, and Yong Zhang.
\newblock Etran: Energy-based transferability estimation.
\newblock In \emph{ICCV}, pages 18613--18622, 2023.

\bibitem[Grill et~al.(2020)Grill, Strub, Altch{\'e}, Tallec, Richemond, Buchatskaya, Doersch, Avila~Pires, Guo, Gheshlaghi~Azar, et~al.]{byol}
Jean-Bastien Grill, Florian Strub, Florent Altch{\'e}, Corentin Tallec, Pierre Richemond, Elena Buchatskaya, Carl Doersch, Bernardo Avila~Pires, Zhaohan Guo, Mohammad Gheshlaghi~Azar, et~al.
\newblock Bootstrap your own latent-a new approach to self-supervised learning.
\newblock In \emph{NeurIPS}, pages 21271--21284, 2020.

\bibitem[Hanin and Nica(2020)]{Finite_WIDTH}
Boris Hanin and Mihai Nica.
\newblock Finite depth and width corrections to the neural tangent kernel.
\newblock In \emph{ICLR}, 2020.

\bibitem[He and Su(2020)]{local}
Hangfeng He and Weijie Su.
\newblock The local elasticity of neural networks.
\newblock In \emph{ICLR}, 2020.

\bibitem[He et~al.(2015)He, Zhang, Ren, and Sun]{kaiming}
Kaiming He, Xiangyu Zhang, Shaoqing Ren, and Jian Sun.
\newblock Delving deep into rectifiers: Surpassing human-level performance on imagenet classification.
\newblock In \emph{ICCV}, 2015.

\bibitem[He et~al.(2016)He, Zhang, Ren, and Sun]{resnet}
Kaiming He, Xiangyu Zhang, Shaoqing Ren, and Jian Sun.
\newblock Deep residual learning for image recognition.
\newblock In \emph{CVPR}, pages 770--778, 2016.

\bibitem[He et~al.(2020)He, Fan, Wu, Xie, and Girshick]{moco-v1}
Kaiming He, Haoqi Fan, Yuxin Wu, Saining Xie, and Ross Girshick.
\newblock Momentum contrast for unsupervised visual representation learning.
\newblock In \emph{CVPR}, pages 9729--9738, 2020.

\bibitem[Huang et~al.(2017)Huang, Liu, Van Der~Maaten, and Weinberger]{densenet}
Gao Huang, Zhuang Liu, Laurens Van Der~Maaten, and Kilian~Q Weinberger.
\newblock Densely connected convolutional networks.
\newblock In \emph{CVPR}, pages 4700--4708, 2017.

\bibitem[Jacot et~al.(2018)Jacot, Gabriel, and Hongler]{NTK}
Arthur Jacot, Franck Gabriel, and Cl{\'e}ment Hongler.
\newblock Neural tangent kernel: Convergence and generalization in neural networks.
\newblock In \emph{NeurIPS}, 2018.

\bibitem[Kendall et~al.(2018)Kendall, Gal, and Cipolla]{multi}
Alex Kendall, Yarin Gal, and Roberto Cipolla.
\newblock Multi-task learning using uncertainty to weigh losses for scene geometry and semantics.
\newblock In \emph{CVPR}, pages 7482--7491, 2018.

\bibitem[Kornblith et~al.(2018)Kornblith, Shlens, and Le]{transfer_better}
Simon Kornblith, Jonathon Shlens, and Quoc~V. Le.
\newblock Do better imagenet models transfer better?
\newblock In \emph{CVPR}, pages 2656--2666, 2018.

\bibitem[Krause et~al.(2013)Krause, Deng, Stark, and Fei-Fei]{cars}
Jonathan Krause, Jia Deng, Michael Stark, and Li Fei-Fei.
\newblock Collecting a large-scale dataset of fine-grained cars.
\newblock 2013.

\bibitem[Krizhevsky et~al.(2009)Krizhevsky, Hinton, et~al.]{cifar}
Alex Krizhevsky, Geoffrey Hinton, et~al.
\newblock Learning multiple layers of features from tiny images.
\newblock 2009.

\bibitem[Li et~al.(2020)Li, Zhou, Xiong, and Hoi]{pcl}
Junnan Li, Pan Zhou, Caiming Xiong, and Steven~CH Hoi.
\newblock Prototypical contrastive learning of unsupervised representations.
\newblock \emph{arXiv}, 2020.

\bibitem[Li et~al.(2023)Li, Hu, Ge, Shan, and Duan]{ped}
Xiaotong Li, Zixuan Hu, Yixiao Ge, Ying Shan, and Ling-Yu Duan.
\newblock Exploring model transferability through the lens of potential energy.
\newblock In \emph{ICCV}, pages 5429--5438, 2023.

\bibitem[Li et~al.(2021)Li, Jia, Sang, Zhu, Green, Wang, and Gong]{nleep}
Yandong Li, Xuhui Jia, Ruoxin Sang, Yukun Zhu, Bradley Green, Liqiang Wang, and Boqing Gong.
\newblock Ranking neural checkpoints.
\newblock In \emph{CVPR}, pages 2663--2673, 2021.

\bibitem[Lin et~al.(2017)Lin, Goyal, Girshick, He, and Doll{\'a}r]{focal}
Tsung-Yi Lin, Priya Goyal, Ross Girshick, Kaiming He, and Piotr Doll{\'a}r.
\newblock Focal loss for dense object detection.
\newblock In \emph{ICCV}, pages 2980--2988, 2017.

\bibitem[Mika et~al.(1999)Mika, Ratsch, Weston, Scholkopf, and Mullers]{LDA}
Sebastian Mika, Gunnar Ratsch, Jason Weston, Bernhard Scholkopf, and Klaus-Robert Mullers.
\newblock Fisher discriminant analysis with kernels.
\newblock In \emph{Neural networks for signal processing IX: Proceedings of the 1999 IEEE signal processing society workshop (cat. no. 98th8468)}, pages 41--48, 1999.

\bibitem[Mohamadi and Sutherland(2022)]{fastntk}
Mohamad~Amin Mohamadi and Danica~J. Sutherland.
\newblock A fast, well-founded approximation to the empirical neural tangent kernel.
\newblock In \emph{ICML}, 2022.

\bibitem[Nguyen et~al.(2020)Nguyen, Hassner, Seeger, and Archambeau]{leep}
Cuong Nguyen, Tal Hassner, Matthias Seeger, and Cedric Archambeau.
\newblock Leep: A new measure to evaluate transferability of learned representations.
\newblock In \emph{ICML}, pages 7294--7305, 2020.

\bibitem[Nilsback and Zisserman(2008)]{flowers}
Maria-Elena Nilsback and Andrew Zisserman.
\newblock Automated flower classification over a large number of classes.
\newblock In \emph{2008 Sixth Indian Conference on Computer Vision, Graphics Image Processing}, pages 722--729. IEEE, 2008.

\bibitem[P{\'a}ndy et~al.(2022)P{\'a}ndy, Agostinelli, Uijlings, Ferrari, and Mensink]{gbc}
Michal P{\'a}ndy, Andrea Agostinelli, Jasper Uijlings, Vittorio Ferrari, and Thomas Mensink.
\newblock Transferability estimation using bhattacharyya class separability.
\newblock In \emph{CVPR}, pages 9172--9182, 2022.

\bibitem[Parkhi et~al.(2012)Parkhi, Vedaldi, Zisserman, and Jawahar]{pets}
Omkar~M Parkhi, Andrea Vedaldi, Andrew Zisserman, and CV Jawahar.
\newblock Cats and dogs.
\newblock In \emph{CVPR}, pages 3498--3505, 2012.

\bibitem[Reynolds et~al.(2009)]{gmm}
Douglas~A Reynolds et~al.
\newblock Gaussian mixture models.
\newblock \emph{Encyclopedia of biometrics}, 741\penalty0 (659-663), 2009.

\bibitem[Sandler et~al.(2018)Sandler, Howard, Zhu, Zhmoginov, and Chen]{mobilenetv2}
Mark Sandler, Andrew Howard, Menglong Zhu, Andrey Zhmoginov, and Liang-Chieh Chen.
\newblock Mobilenetv2: Inverted residuals and linear bottlenecks.
\newblock In \emph{CVPR}, pages 4510--4520, 2018.

\bibitem[Shao et~al.(2022)Shao, Zhao, Ge, Zhang, Yang, Wang, Shan, and Luo]{sfda}
Wenqi Shao, Xun Zhao, Yixiao Ge, Zhaoyang Zhang, Lei Yang, Xiaogang Wang, Ying Shan, and Ping Luo.
\newblock Not all models are equal: Predicting model transferability in a self-challenging fisher space.
\newblock In \emph{ECCV}, pages 286--302, 2022.

\bibitem[Shieh(1998)]{kendall}
Grace~S. Shieh.
\newblock A weighted kendall's tau statistic.
\newblock \emph{Statistics Probability Letters}, 39\penalty0 (1):\penalty0 17--24, 1998.

\bibitem[Szegedy et~al.(2015)Szegedy, Liu, Jia, Sermanet, Reed, Anguelov, Erhan, Vanhoucke, and Rabinovich]{googlenet}
Christian Szegedy, Wei Liu, Yangqing Jia, Pierre Sermanet, Scott Reed, Dragomir Anguelov, Dumitru Erhan, Vincent Vanhoucke, and Andrew Rabinovich.
\newblock Going deeper with convolutions.
\newblock In \emph{CVPR}, pages 1--9, 2015.

\bibitem[Szegedy et~al.(2016)Szegedy, Vanhoucke, Ioffe, Shlens, and Wojna]{inceptionv3}
Christian Szegedy, Vincent Vanhoucke, Sergey Ioffe, Jon Shlens, and Zbigniew Wojna.
\newblock Rethinking the inception architecture for computer vision.
\newblock In \emph{CVPR}, pages 2818--2826, 2016.

\bibitem[Tan et~al.(2019)Tan, Chen, Pang, Vasudevan, Sandler, Howard, and Le]{mnasnet}
Mingxing Tan, Bo Chen, Ruoming Pang, Vijay Vasudevan, Mark Sandler, Andrew Howard, and Quoc~V Le.
\newblock Mnasnet: Platform-aware neural architecture search for mobile.
\newblock In \emph{CVPR}, pages 2820--2828, 2019.

\bibitem[Tian et~al.(2020)Tian, Sun, Poole, Krishnan, Schmid, and Isola]{infomin}
Yonglong Tian, Chen Sun, Ben Poole, Dilip Krishnan, Cordelia Schmid, and Phillip Isola.
\newblock What makes for good views for contrastive learning?
\newblock In \emph{NeurIPS}, pages 6827--6839, 2020.

\bibitem[Tran et~al.(2019)Tran, Nguyen, and Hassner]{NCE}
Anh~T Tran, Cuong~V Nguyen, and Tal Hassner.
\newblock Transferability and hardness of supervised classification tasks.
\newblock In \emph{ICCV}, pages 1395--1405, 2019.

\bibitem[Wang et~al.(2022)Wang, Yu, and Perdikaris]{pinn}
Sifan Wang, Xinling Yu, and Paris Perdikaris.
\newblock When and why pinns fail to train: A neural tangent kernel perspective.
\newblock \emph{Journal of Computational Physics}, 449:\penalty0 110768, 2022.

\bibitem[Weston et~al.(1999)Weston, Watkins, et~al.]{SVM}
Jason Weston, Chris Watkins, et~al.
\newblock Support vector machines for multi-class pattern recognition.
\newblock In \emph{Esann}, pages 219--224, 1999.

\bibitem[Wu et~al.(2018)Wu, Xiong, Yu, and Lin]{indis}
Zhirong Wu, Yuanjun Xiong, Stella~X Yu, and Dahua Lin.
\newblock Unsupervised feature learning via non-parametric instance discrimination.
\newblock In \emph{CVPR}, pages 3733--3742, 2018.

\bibitem[Xiao et~al.(2010)Xiao, Hays, Ehinger, Oliva, and Torralba]{sun397}
Jianxiong Xiao, James Hays, Krista~A Ehinger, Aude Oliva, and Antonio Torralba.
\newblock Sun database: Large-scale scene recognition from abbey to zoo.
\newblock In \emph{CVPR}, pages 3485--3492. IEEE, 2010.

\bibitem[You et~al.(2021)You, Liu, Wang, and Long]{logme}
Kaichao You, Yong Liu, Jianmin Wang, and Mingsheng Long.
\newblock Logme: Practical assessment of pre-trained models for transfer learning.
\newblock In \emph{ICML}, pages 12133--12143, 2021.

\bibitem[Y{\"u}ce et~al.(2022)Y{\"u}ce, Ortiz-Jim{\'e}nez, Besbinar, and Frossard]{structured}
Gizem Y{\"u}ce, Guillermo Ortiz-Jim{\'e}nez, Beril Besbinar, and Pascal Frossard.
\newblock A structured dictionary perspective on implicit neural representations.
\newblock In \emph{CVPR}, pages 19228--19238, 2022.

\end{thebibliography}
}

\maketitlesupplementary

\appendix

In this supplementary material, we first present theoretical proof of our approach. Next, we introduce implementation details about LEAD and conduct more ablation studies for our approach to validate its effectiveness. Finally, we provide performance comparisons under different measurements and fine-tuning results of the ground-truth.

\section{Theoretical Proof}
\label{Theoretical proof}
Below, we will provide detailed proof of the theoretical results presented in the methodology section.

\noindent \textbf{Notation.} First, We recall the notation that we used in the main paper as well as this appendix:

\noindent $\{\mathcal{X}, \mathcal{Y}\}$ denotes the downstream dataset. $\mathcal{F}$ denotes the logit function. $\eta$ denotes the learning rate. $l$ denotes root mean square of network widths. $\Phi$ denotes NTK matrix when $l$ approaching infinite. $\mathbb{I}$ denotes the identity matrix. To simplify the notation, $\cdot$ denotes both dot product and scalar multiplication, $e^\cdot, \cdot^2$ denotes both scalar and matrix power operations, and $\frac{\mathrm{d}\times}{\mathrm{d}\cdot}$ denotes both differentiation of scalar functions and the Jacobian matrix of vector functions.

\noindent\textbf{Dynamical Equation.} Subsequently, we provide the proof for obtaining the continuous dynamical equation in Eq. (5) of the main paper through the limit approximation method.
    \vspace{-0.15cm}
\begin{conclusion}
Consider an infinite-width neural network $\mathcal{F}$ and a downstream dataset $\mathcal{T}=\{\mathcal{X}, \mathcal{Y}\}$. The evolution process of the output logits $\mathcal{F}_t(\mathcal{X})$ will follow the following differential equation:
\vspace{-0.05cm}
\begin{equation}
\label{eq:ode_appendix}
\frac{\mathrm{d}\mathcal{F}_t}{\mathrm{d} t}   =-\eta\cdot \Phi\cdot \frac{\mathrm{d}\mathcal{L}_t}{\mathrm{d}\mathcal{F}_t}, \; \mathcal{F}_0 = log_{init}.
\end{equation}
\end{conclusion}
\vspace{-0.3cm}
\begin{proof}
We need to consider the variation trend of $\mathcal{F}$. Therefore, we calculate its derivative at time $t$ according to the definition:
\vspace{-0.06cm}
\begin{equation}
\label{eq:1}
    \begin{aligned}
\frac{\mathrm{d} \mathcal{F}_{t}(\mathcal{X})}{\mathrm{d} t} & =\lim _{\Delta t \rightarrow 0} \frac{\mathcal{F}_{t+\Delta t}(\mathcal{X})-\mathcal{F}_t(\mathcal{X})}{\Delta t} \\
& =\lim _{\Delta t \rightarrow 0} \frac{\mathcal{F}\left(\mathcal{X}, \theta_{t+\Delta t}\right)-\mathcal{F}\left(\mathcal{X}, \theta_t\right)}{\Delta t}.
\end{aligned}
\end{equation}

\noindent Through Taylor's theorem, we obtain the asymptotic series decomposition of $\mathcal{F}$ concerning the variation of $\theta$:
\vspace{-0.25cm}

\begin{equation}
\label{eq:2}
    \begin{aligned}
\mathcal{F}\left(\mathcal{X}, \theta_{t+\Delta t}\right) & =\mathcal{F}\left(\mathcal{X}, \theta_t\right)+\frac{\mathrm{d} \mathcal{F}\left(\mathcal{X}, \theta_t\right)}{\mathrm{d} \theta}\left(\theta_{t+\Delta t}-\theta_t\right) \\
& +O\left(\left(\theta_{t+\Delta t}-\theta_t\right)^2\right),
\end{aligned}
\end{equation}

\noindent where $O(\cdot)$ denotes a infinitesimal quantity which has the same order with $\cdot$, describing the remainder of the asymptotic series. Meanwhile, according to the optimization of the gradient descent, we can determine the equivalent effects in corresponding continuous time:
\vspace{-0.3cm}

\begin{equation}
    \label{eq:3}
    \begin{aligned}
    \theta_{t+\Delta t}-\theta_t=-\eta& \cdot \frac{\mathrm{d} \mathcal{L}\left(\mathcal{F}\left(\mathcal{X}, \theta_t\right), \mathcal{Y}\right)}{\mathrm{d} \theta} \cdot \Delta t, \\
    O\left(\left(\theta_{t+\Delta t}-\theta_t\right)^2\right) &= O\left(\eta^2\cdot\frac{\mathrm{d} \mathcal{L}_t}{\mathrm{d} \theta}\cdot\frac{\mathrm{d} \mathcal{L}_t}{\mathrm{d} \theta}{\cdot\Delta t^2}\right) \\
    &=O\left(\left(\Delta t\right)^2\right).
    \end{aligned}
\end{equation}
Combining Eq. (\ref{eq:2}) and Eq. (\ref{eq:3}), we obtain the asymptotic series decomposition of $\mathcal{F}$ concerning the variation of $t$:

\vspace{-0.1cm}
\begin{equation}
    \label{eq:4}
\begin{aligned}
& \mathcal{F}\left(\mathcal{X}, \theta_{t+\Delta t}\right)-\mathcal{F}\left(\mathcal{X}, \theta_t\right) \\
=\; & \frac{\mathrm{d} \mathcal{F}\left(\mathcal{X}, \theta_t\right)}{\mathrm{d} \theta}\left(\theta_{t+\Delta t}-\theta_t\right)+O\left(\left(\theta_{t+\Delta t}-\theta_t\right)^2\right) \\
= & -\eta \cdot \frac{\mathrm{d} \mathcal{F}\left(\mathcal{X}, \theta_t\right)}{\mathrm{d} \theta} \cdot \frac{\mathrm{d} \mathcal{L}\left(\mathcal{F}\left(\mathcal{X}, \theta_t\right), \mathcal{Y}\right)}{\mathrm{d} \theta} \Delta t+O\left(\left(\Delta t\right)^2\right).
\end{aligned}
\end{equation}
Combining Eq. (\ref{eq:1}) and Eq. (\ref{eq:4}) and employing the property of $O(\cdot)$, we can use limit approximation, discarding the remainder terms with a limit value of 0, to determine the value of the derivative:

\vspace{-0.1cm}
\begin{equation}
    \label{eq:5}
    \begin{aligned}
\frac{\mathrm{d} \mathcal{F}_t(\mathcal{X})}{\mathrm{d} t} & =\lim _{\Delta t \rightarrow 0}-\eta \cdot \frac{\mathrm{d} \mathcal{F}\left(\mathcal{X}, \theta_t\right)}{\mathrm{d} \theta} \cdot \frac{\mathrm{d} \mathcal{L}\left(\mathcal{F}\left(\mathcal{X}, \theta_t\right), \mathcal{Y}\right)}{\mathrm{d} \theta} \\
& + O\left(\left(\Delta_t\right)^2\right) / \Delta t \\
& =-\eta \cdot \frac{\mathrm{d} \mathcal{F}\left(\mathcal{X}, \theta_t\right)}{\mathrm{d} \theta} \cdot \frac{\mathrm{d} \mathcal{L}\left(\mathcal{F}\left(\mathcal{X}, \theta_t\right), \mathcal{Y}\right)}{\mathrm{d} \theta} \\
& =-\eta \cdot \frac{\mathrm{d} \mathcal{F}_t}{\mathrm{d} \theta} \cdot \frac{\mathrm{d} \mathcal{L}_t}{\mathrm{d} \theta}.
\end{aligned}
\end{equation}
where the last line simplifies the notation for convenience.

\noindent Through the chain rule, we have:
\begin{equation}
    \label{eq:6}
    \frac{\mathrm{d} \mathcal{F}_t}{\mathrm{d} t}=-\eta \cdot \frac{\mathrm{d} \mathcal{L}_t}{\mathrm{d} \mathcal{F}_t} \cdot \left(\frac{\mathrm{d} \mathcal{F}_t}{\mathrm{d} \theta}\cdot \frac{\mathrm{d} \mathcal{F}_t}{\mathrm{d} \theta} \right).
\end{equation}

\noindent Finally, through the definition and the constant-preserving property of the NTK \cite{NTK}, we obtain the dynamical equation with its initial value condition:
\qedhere
\begin{equation}
    \label{eq:7}
    \frac{\mathrm{d}\mathcal{F}_t}{\mathrm{d} t}   =-\eta\cdot \Phi\cdot \frac{\mathrm{d}\mathcal{L}_t}{\mathrm{d}\mathcal{F}_t}, \; \mathcal{F}_0 = log_{init}.
\end{equation}
\end{proof}

\noindent\textbf{Closed-form Solution.} Finally, we solve the ordinary differential equation (ODE) in Eq. (\ref{eq:7}) through separation of variables and integration along the time dimension. As a commonly used scene in theoretical analysis, we consider the case where $\mathcal{L}$ is Mean Squared Error loss. And we can obtain concise closed-form solution as shown in Eq. (6) of the main paper:

\vspace{-0.1cm}
\begin{conclusion}
    Consider the case where $\mathcal{L}$ is MSE loss. The solution of logits $\mathcal{F}_t(\mathcal{X})$ to the equation presented in Eq. (\ref{eq:7}) can be expressed as follows:
    \vspace{-0.1cm}
\begin{equation}
\label{eq:solution_appendix}
\mathbb{E}\left(\mathcal{F}_t\left(\mathcal{X}\right)\right)=\left(\mathbb{I}-e^{-\eta \Phi\cdot t}\right) \mathcal{Y}+e^{-\eta \Phi\cdot t} \cdot log_{init}.
\end{equation}
\end{conclusion}

\begin{proof}
Due to $\frac{\mathrm{d} \mathcal{L}_t}{\mathrm{d} \mathcal{F}_t}=\mathcal{F}_t-\mathcal{Y}$, we have:

\begin{equation}
    \label{eq:8}
    \frac{\mathrm{d} \mathcal{F}_t}{\mathrm{d} t}=-\eta \cdot \Phi \cdot\left(\mathcal{F}_t-\mathcal{Y}\right), \; \mathcal{F}_0 = log_{init}.
\end{equation}

\vspace{-0.1cm}
\noindent In order to transform both sides of Eq. (\ref{eq:8}) into independent differentials, we first introduce an integrating factor $e^{\eta\Phi t}$:

\vspace{-0.3cm}
\begin{equation}
\label{eq:9}
\begin{aligned}
& e^{\eta \Phi t}\; \mathrm{d} \mathcal{F}_t=-\eta \Phi \cdot e^{\eta \Phi t}\left(\mathcal{F}_t-\mathcal{Y}\right) \mathrm{d} t \\
 \Leftrightarrow\; &e^{\eta \Phi t} \;\mathrm{d} \mathcal{F}_t+\eta \Phi \cdot e^{\eta \Phi t} \mathcal{F}_t \;\mathrm{d} t=\eta \Phi \cdot e^{\eta \Phi t}\cdot \mathcal{Y}\; \mathrm{d} t.
\end{aligned}
\end{equation}

\vspace{-0.1cm}
\noindent Through the differential properties of composite functions, the left-hand side of Eq. (\ref{eq:9}) is equivalent to the differential of a composite function:

\vspace{-0.08cm}
\begin{equation}
    \label{eq:10}
    \begin{aligned}
& e^{\eta \Phi t} \;\mathrm{d} \mathcal{F}_t+\eta \Phi \cdot e^{\eta \Phi t} \mathcal{F}_t \;\mathrm{d} t \\
=\;& e^{\eta \Phi t}\; \mathrm{d} \mathcal{F}_t+\mathcal{F}_t \; \mathrm{d} e^{\eta \Phi t} = \mathrm{d}\left(\mathcal{F}_t \cdot e^{n \Phi t}\right).
\end{aligned}
\end{equation}

\vspace{-0.1cm}
\noindent Combining Eq. (\ref{eq:9}) and Eq. (\ref{eq:10}), we get the following equation whose left and right sides are determined by the differential of independent composite functions:

\vspace{-0.1cm}
\begin{equation}
    \label{eq:11}
    \mathrm{d}\left(\mathcal{F}_t \cdot e^{n \Phi t}\right) = \mathcal{Y}\; \mathrm{d}\left( e^{\eta \Phi t}\right).
\end{equation}
By integrating over the time dimension $t$, we have:

\vspace{-0.1cm}
\begin{equation}
    \label{eq:12}
    \begin{aligned}
    &\mathcal{F}_t \cdot e^{n \Phi t}|_{0}^{t} = \mathcal{Y}\cdot e^{\eta \Phi t}|_{0}^{t} \\
    \Leftrightarrow\; &\mathcal{F}_t \cdot e^{n \Phi t} - \mathcal{F}_0 = \mathcal{Y}\cdot (e^{\eta \Phi t} - \mathbb{I}).
    \end{aligned}    
\end{equation}
Substituting the initial condition in Eq. (\ref{eq:8}), we obtain the solution to the dynamical equation:

\vspace{-0.3cm}
\begin{equation}
    \label{eq:13}
\mathcal{F}_t\left(\mathcal{X}\right)=\left(\mathbb{I}-e^{-\eta \Phi\cdot t}\right) \mathcal{Y}+e^{-\eta \Phi\cdot t} \cdot log_{init}. 
\end{equation}
Since the constant-preserving property of $\Phi$ requires taking the expectation over the randomly initialized part of $\mathcal{F}$, our results need to take the expectation as well:
\begin{equation}
\label{eq:solution}
\mathbb{E}\left(\mathcal{F}_t\left(\mathcal{X}\right)\right)=\left(\mathbb{I}-e^{-\eta \Phi\cdot t}\right) \mathcal{Y}+e^{-\eta \Phi\cdot t} \cdot log_{init}. \qedhere
\end{equation}
\end{proof}

\vspace{-0.4cm}
\section{Implementation Details}
\vspace{-0.1cm}
The implementation of our method are presented in the section of methodology and experiments. Additionally, we present more detailed information and the pseudocode of the overall algorithm in this section.

\noindent\textbf{Classification Algorithm.} 
In the experiments and ablation study section of the main paper, we have explained the reasons for choosing the SVM algorithm as classification algorithm to obtain $log_{init}$. Here, we will provide a detailed introduction to the Multi-class SVM. The original SVM is a binary classifier, that can only handle two classes problems. However, for multiclass problems, we can utilize the One-vs-Rest (OvR) strategy \cite{SVM}. For a problem with K classes, we train K different binary classifiers, with each classifier specifically addressing one class, treating it as the positive class, and considering all other classes as negative. For a given sample, predictions are made using these classifiers and the prediction values are normalized to obtain the predicted probabilities (\textit{i.e.,} logits) for K classes.

\noindent\textbf{Calculation of NTK.} 
For the computation of the NTK, we employ the approximation algorithm proposed in \cite{fastntk}. Specifically, we concatenate the model backbone with the classification head, a randomly initialized MLP (with two hidden layers of width 1024 and 2048). The combination of these two parts constitutes $\mathcal{F}$ to output logits. Subsequently, we calculate the NTK using the following equation:

\vspace{-0.2cm}
\begin{equation}
    \label{eq:fastntk}
    \Phi_{i,j}(\mathcal{X}, \mathcal{X}) = \left[\nabla_\theta  \sum_{k=1}^K \mathcal{F}^{(k)}\left(x_i, \theta\right)\right]\left[\nabla_\theta  \sum_{k=1}^K \mathcal{F}^{(k)}\left(x_j, \theta\right)\right]
\end{equation}
where $\Phi_{i,j}$ denotes the element of i-th row and j-th column in $\Phi$. $x_i$ denotes i-th sample of $\mathcal{X}$. $K$ denotes the output dimension of $\mathcal{F}$, $\mathcal{F}^{(k)}$ denotes the k-th output dimension and $\nabla$ is only utilized on the MLP for efficiency. The computation method in Eq. (\ref{eq:fastntk}) reduces the computational cost of the NTK. Meanwhile, \cite{fastntk} provides both theoretical proof and experimental validation that the eigenvalue range obtained through this approximation is close to that of the original computation method in \cite{NTK}.

\noindent\textbf{Pseudocode.}
We provide the pseudocode for our LEAD to demonstrate the complete computation procedure.

\vspace{-0.2cm}
\begin{algorithm}
\footnotesize
\caption{The algorithm of our method LEAD}
\KwIn{
Downstream dataset $\{\mathcal{X},\mathcal{Y}\}$, with $K$ classes; Model zoo $\{\Psi_i\}_{i=1}^{m}$ with their backbone $\{f_i\}_{i=1}^{m}$; The sample size hyper-meter $S$ for class-aware decomposition; The time hyper-parameter $t$.
}
\KwOut{The transferability score $\{P_i\}_{i=1}^m$ for each model in the model zoo and the predicted rank.}
\For{$\Psi_i$ in Model zoo}{
    \,\, Encode images $\mathcal{X}$ to feature embeddings and feed features and corresponding labels into the classification algorithm $\mathcal{C}$ to obtain initial state logits: \\
     $\qquad\qquad\qquad \mathcal{Z}=f_i(\mathcal{X})$, $\; log_{init} = \mathcal{C}(\mathcal{Z}, \mathcal{Y});$ \\
    \,\, Concatenate a randomly initialized MLP $h$ as a classifica- tion head after its backbone $f$ and combine them as logit function $\mathcal{F}=h\circ f$. \\
    \For{k in range(K)}{
    \,\, Select $S$ samples $\mathcal{X}_k:=\{x_{k_i}\}_{i=1}^{S}$ from k-th class. \\
    \,\, Compute the NTK matrix for this class: \\
    $\qquad \Phi_{u,v}(\mathcal{X}_k, \mathcal{X}_k)=\left[\nabla_\theta  \sum_{j=1}^K \mathcal{F}^{(j)}\left(x_{k_u}, \theta\right)\right]\cdot$ \\
    $\qquad\qquad\qquad\qquad\quad\, \left[\nabla_\theta  \sum_{j=1}^K \mathcal{F}^{(j)}\left(x_{k_v}, \theta\right)\right]$; \\
    \,\, for $1\leq u, v\leq K$. \\
    \,\, Employ eigenvalue decomposition to $\Phi(\mathcal{X}_k, \mathcal{X}_k)$ and obtain the mean of eigenvalues $\overline{\lambda}$. \\
    \,\, We can get the prediction value of the final state logits of samples belonging to this class:
    $\mathbb{E}\left(\mathcal{F}_t\left(x\right)\right)=\left(\mathbb{I}-e^{-\overline{\lambda}\cdot t}\right) y+e^{-\overline{\lambda}\cdot t} \cdot log_{init}(x);$ \\
    \,\, for $x$ belonging to the k-th class.
    } 
    \,\, Feed the predicted final state logits $\mathcal{F}_t\left(\mathcal{X}\right)$ and ground- truth label $\mathcal{Y}$ into the Cross-Entropy loss to obtain the score $P_i$ for the model $\Psi_i$.
    
}
Rank models $\{\Psi_i\}_{i=1}^{m}$ according to their scores $\{P_i\}_{i=1}^m$.
\end{algorithm}

\section{More Ablation Study and Visualization}

\subsection{Initial State Observation}
\begin{figure}[h]
\begin{center}
\includegraphics[width=0.95\linewidth]{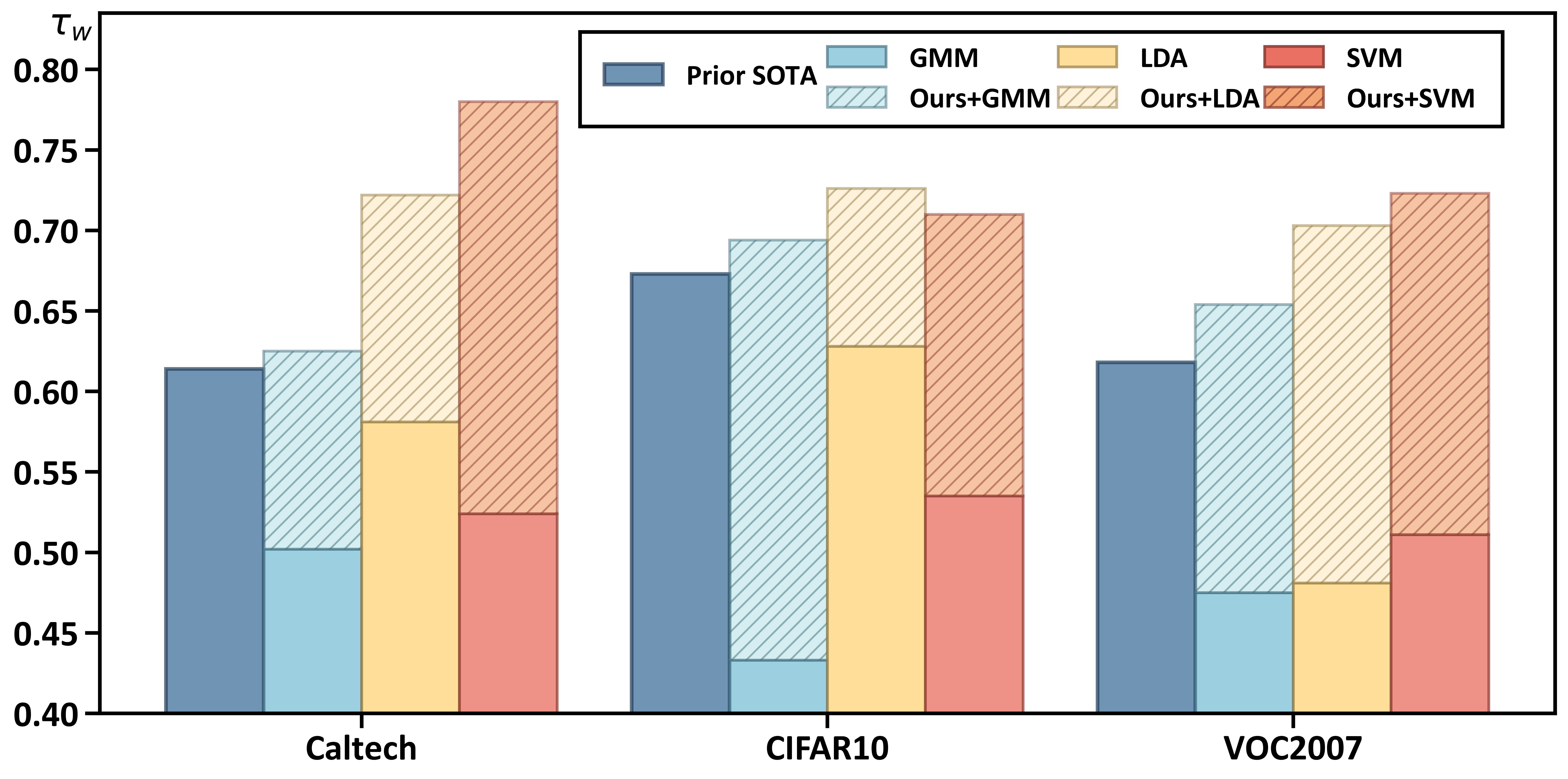}
\setlength{\belowcaptionskip}{-27pt} 
\setlength{\abovecaptionskip}{-1pt} 
\caption{\small Performance comparison of our method under different classifier selection to compute $log_{init}$.}
\label{fig:ml_ablation}
\end{center}
\end{figure}

To capture the observation of $log_{init}$, we feed features and labels of downstream datasets into a classifier that can output the probability of each class. We evaluate several commonly used machine learning algorithms SVM \cite{SVM}, LDA \cite{LDA}, and GMM \cite{gmm}, comparing their impact on performance. As shown in Fig. \ref{fig:ml_ablation}, it's noteworthy that, whatever the classifier employed, direct utilization of $log_{init}$ fails to yield satisfactory results due to overlooking the training dynamics. Conversely, the application of LEAD consistently improves all results to exceed the prior SOTA. This demonstrates that LEAD does not rely on a specific classifier and has uniform applicability to different classifier selections.

\subsection{Different Interpolation Strategies}
\begin{figure}[h]
    \centering
    \includegraphics[width=\linewidth]{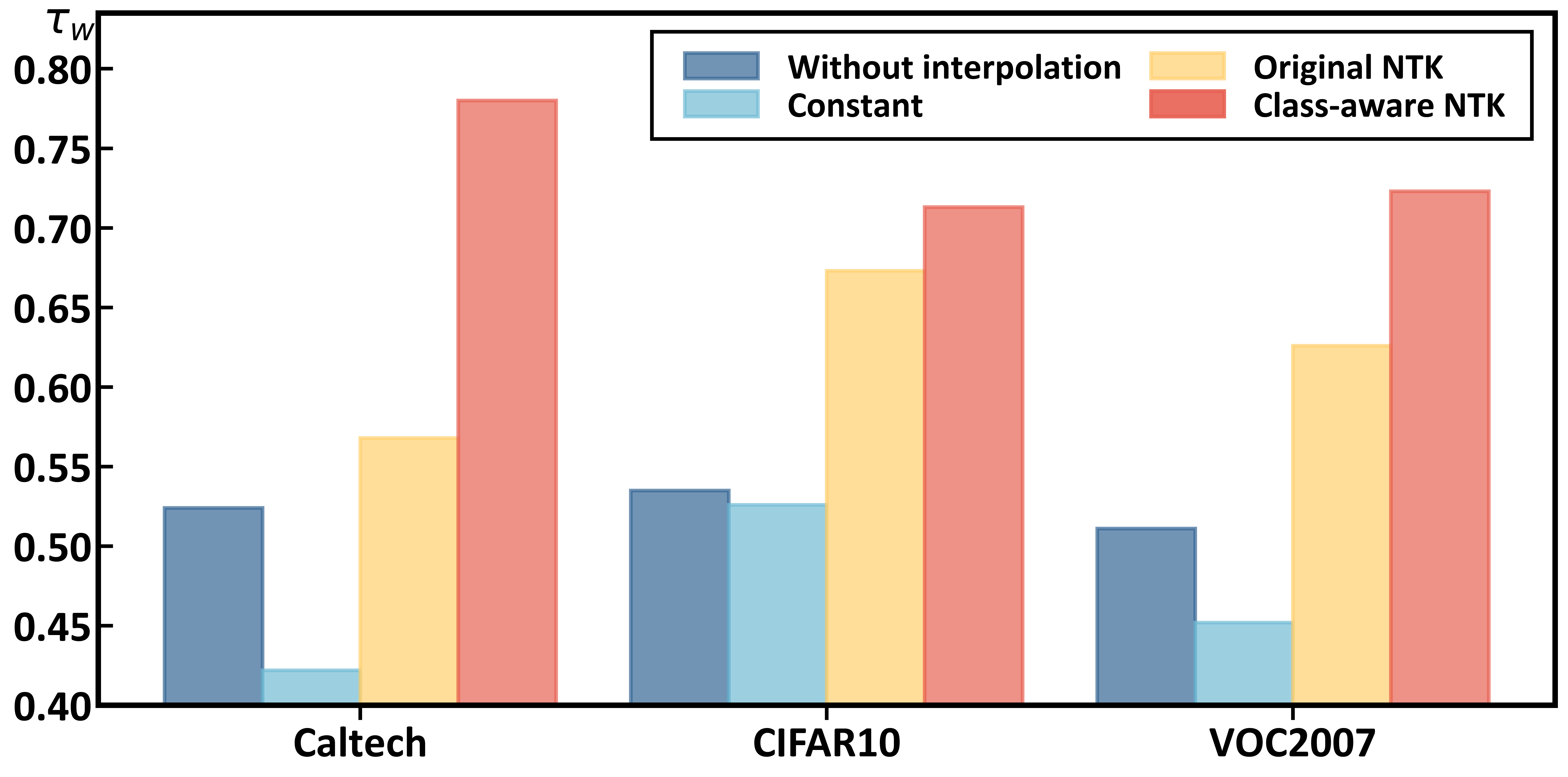}
    \caption{\small Performance comparison under different interpolation strategies.}
    \label{fig:interpo}
\end{figure}

In our method, we use eigenvalues of the class-aware NTK matrix to determine the interpolation coefficient. In this section, we implement other strategies for comparison in Fig. \ref{fig:interpo}. Specifically, `Constant' denotes directly choosing a constant $c$ as the coefficient for all samples and its results are optimal ones selected through grid search within $c\in\{0.1, 0.3, 0.5, 0.7, 0.9\}$. `Original NTK' denotes the computation of the NTK by mixing samples randomly chosen from all classes to determine interpolation coefficients. `Class-aware NTK' denotes the approach we propose, which involves computing NTK separately for each class to determine interpolation coefficients.

As we can see, directly employing a constant will harm the performance because it incorrectly assumes that evolution processes of all samples are the same. `Original NTK', which computes NTK by mixing samples, shows some improvement, especially for datasets with fewer classes like CIFAR10. However, for datasets with a larger number of classes like Caltech, it exhibits only marginal improvement due to a lack of ability to model the evolution of different classes. The results highlight the advantage of our proposed class-aware method in modeling the evolution process.

\vspace{-0.3cm}
\subsection{Network Width}
\begin{figure}[h]
    \centering
    \includegraphics[width=\linewidth]{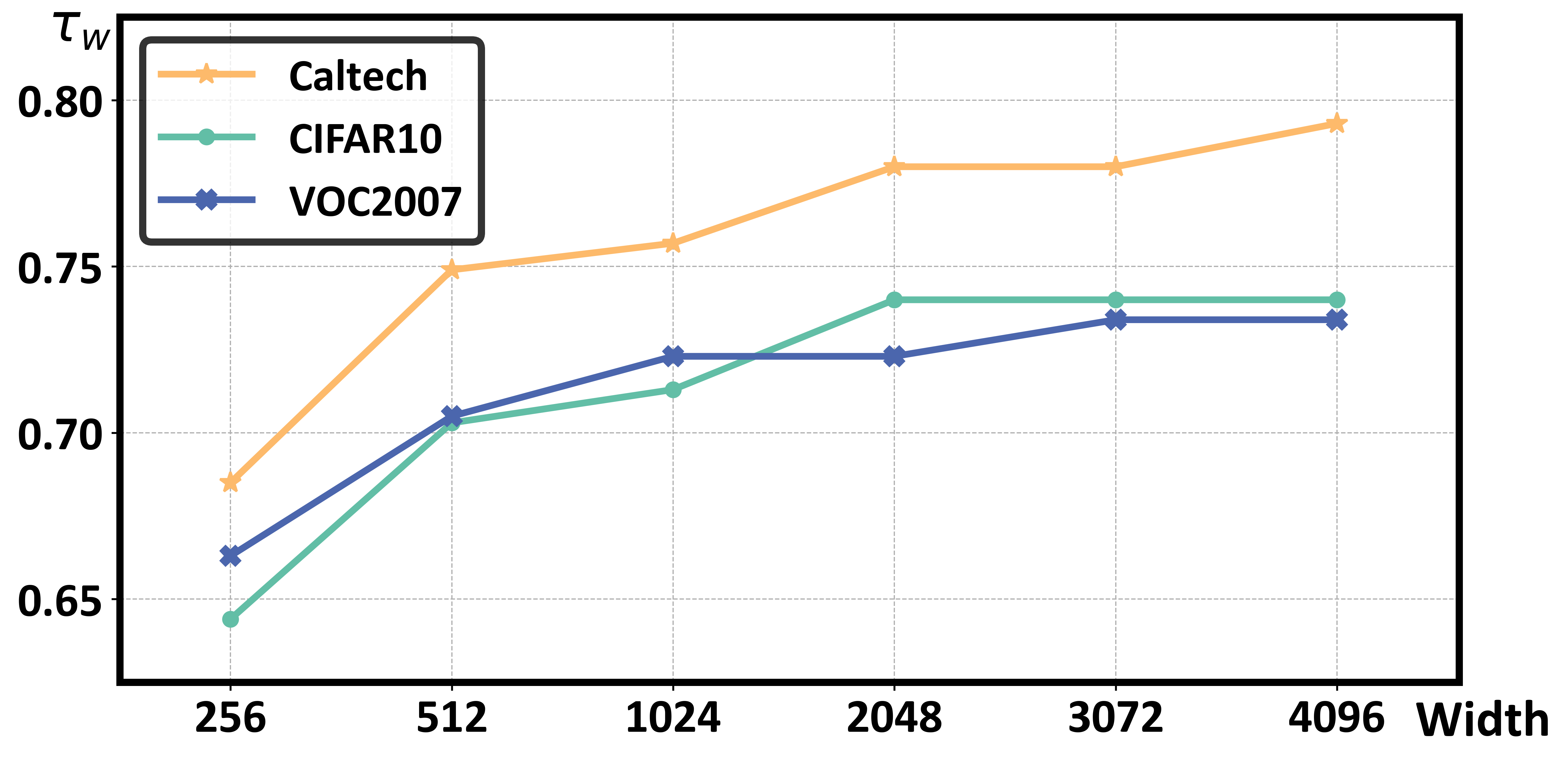}
    \caption{\small Performance comparison under different MLP hidden layer width.}
    \label{fig:width}
\end{figure}

The width of the model backbone is fixed, but the hidden layer width of the classification head MLP can be adjusted. We evaluate the impact of different hidden layer widths on performance, and for the convenience of comparison, we set both two hidden layers to the same width, as shown in Fig. \ref{fig:width}. As we can see, with the increase in the width of the hidden layers, the accuracy of the rank also improves. It indicates that, as the network width increases, the calculation condition of the NTK matrix gradually approach the required theoretical assumption, leading to the improvement of prediction precision. To balance computation cost and performance, we set the widths of two hidden layers to be 1024 and 2048.

\subsection{Logit Predictions for Different Classes}
\begin{figure*}[h]
    \centering
    \includegraphics[width=0.95\linewidth]{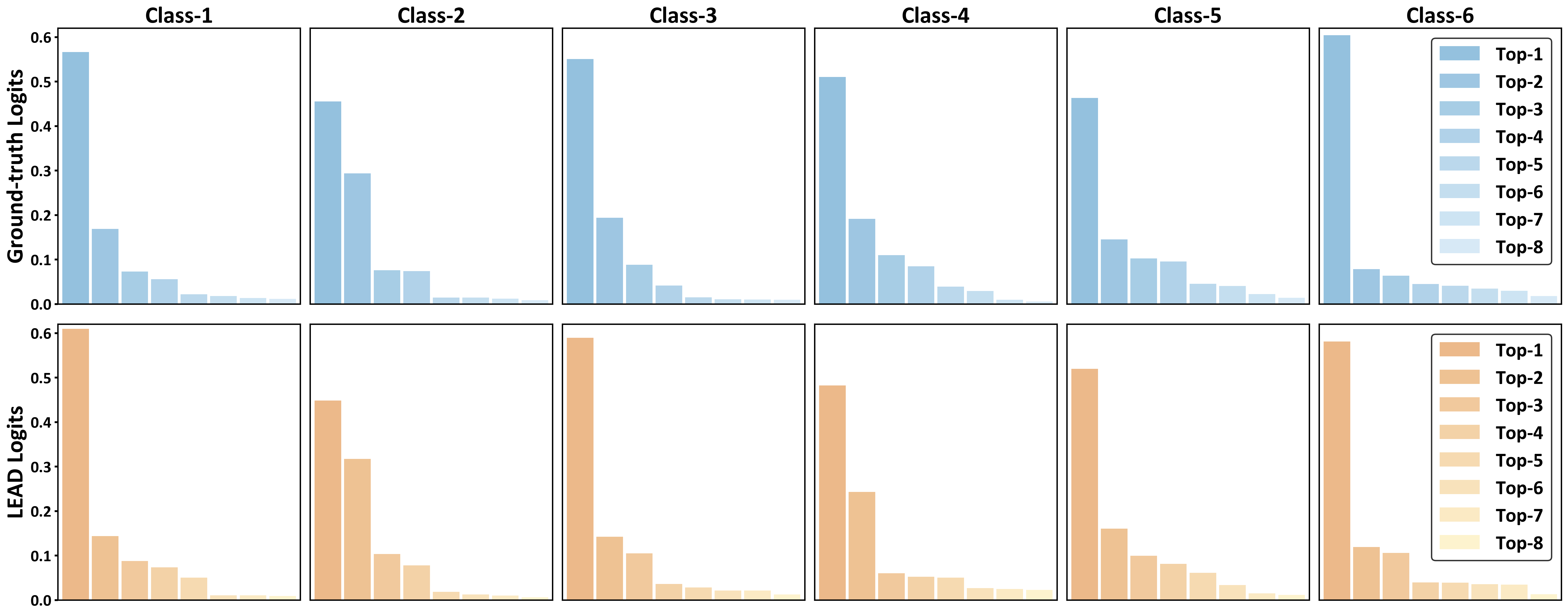}
    \caption{\small Comparison of the average prediction for final state logits between LEAD and Ground-truth results (obtain after fully fine-tuning) on six classes of VOC2007 dataset.}
    \label{fig:logits_class}
\end{figure*}

In Fig. 3 and Fig. 6 of the main paper, we have demonstrated the accuracy of LEAD in predicting logits for the dataset through comparison with prior arts. In this section, we provide visualization results for more detailed comparison. As shown in Fig. \ref{fig:logits_class}, we present the differences between the logits predicted by LEAD and the actual values obtained after fine-tuning, focusing on six classes of the VOC2007 dataset. We can observe that the distribution of predicted logits closely aligns with the actual results across six categories, further highlighting the precision of our dynamical equation-based approach in capturing logit evolution for accurate ranking sequences.

\section{Different Measurements of Transferability}
Apart from utilizing weighted Kendall's tau ($\tau_w$), here we employ various measurement metrics of the rank correlation to assess the performance of our method LEAD. These evaluation metrices include Kendall's tau ($\tau$), weighted Pearson's correlation \,($r_w$), and top-k relative accuracy\, (Rel-k).

\begin{table}[h]
\caption{\small Performance comparison under different measurement metrics of rank correlation assessment on Caltech \cite{caltech101}, CIFAR10 \cite{cifar}, and VOC2007 \cite{voc2007} datasets using self-supervised model zoo.}
\label{table:different_mea}
\resizebox{\linewidth}{!}{
\begin{tabular}{c|c|ccccc}
\toprule[1pt]
Dataset                  & Method  & $\tau_w$       & $\tau$         & $r_w$          & Rel-1          & Rel-3      \\ \hline
\multirow{5}{*}{Caltech}  & PACTran \cite{pac} & 0.622          & 0.455          & 0.457          & \textbf{1.000}     & \textbf{1.000} \\
                         & SFDA \cite{sfda}    & 0.523          & 0.424          & 0.627          & 0.995          & \textbf{1.000} \\
                         & ETran \cite{etran}   & 0.405          & 0.455          & 0.410           & 0.990          & 0.992      \\
                         & PED \cite{ped}    & 0.614          & 0.530           & 0.675          & 0.995          & \textbf{1.000} \\
                         & LEAD (Ours)    & \textbf{0.780}  & \textbf{0.758} & \textbf{0.851} & \textbf{1.000}     & \textbf{1.000} \\ \hline
\multirow{5}{*}{CIFAR10} & PACTran \cite{pac}& 0.477          & 0.485          & 0.497          & \textbf{1.000}     & \textbf{1.000} \\
                         & SFDA \cite{sfda}   & 0.619          & 0.545          & 0.558          & \textbf{1.000}     & \textbf{1.000} \\
                         & ETran \cite{etran}  & 0.606          & 0.515          & 0.422          & 0.999          & \textbf{1.000} \\
                         & PED \cite{ped}    & 0.673          & 0.606          & 0.645          & \textbf{1.000}     & \textbf{1.000} \\
                         & LEAD (Ours)   & \textbf{0.713} & \textbf{0.667} & \textbf{0.792} & \textbf{1.000}     & \textbf{1.000} \\ \hline
\multirow{5}{*}{VOC2007} & PACTran \cite{pac} & 0.620           & 0.485          & 0.413          & 0.995          & \textbf{1.000} \\
                         & SFDA \cite{sfda}   & 0.568          & 0.424          & 0.636          & 0.998          & 0.998      \\
                         & ETran \cite{etran}  & 0.376          & 0.455          & 0.468          & 0.974          & 0.998      \\
                         & PED \cite{ped}    & 0.583          & 0.484          & 0.708          & 0.998          & \textbf{1.000} \\
                         & LEAD (Ours)   & \textbf{0.723} & \textbf{0.788} & \textbf{0.847} & \textbf{0.998} & \textbf{1.000} \\
\bottomrule[1pt]
\end{tabular}
}
\end{table}

\noindent Rel-k represents the ratio between the optimal fine-tuning accuracy within the models ranked in the top-k and the best fine-tuning accuracy across all models. To evaluate the robustness of transferability metrics across different measurements, we conduct experiments using self-supervised CNN models on Caltech, CIFAR10, and VOC2007 datasets, as shown in Tab. \ref{table:different_mea}. The results show that $\tau_w$ is highly correlated with other metrics, and it can assign larger weights to models in higher ranking, which we are concerned with. Therefore, following \cite{ped,sfda,gbc}, we employ $\tau_w$ in the main paper to evaluate performance. Meanwhile, our LEAD consistently outperforms prior arts such as PACTran, SFDA, ETran, and PED across the aforementioned measurements, highlighting the superior performance of LEAD. 

\begin{table*}[t]
\caption{The ground-truth results of the 12 supervised pre-trained models on 10 downstream tasks.}
\centering
\label{tab:gt_s}
{%
\begin{tabular}{ccccccccccc}
\toprule[1pt]
\multicolumn{1}{c}{Supervised} & Food                      & Caltech                   & Flowers                   & Cars                      & CIFAR100                  & DTD                       & CIFAR10                   & Pets                      & SUN397                    & VOC2007                \\ \hline
ResNet-34 \cite{resnet}    & 81.99                     & 91.15                     & 95.20                      & 88.63                     & 81.94                     & 72.96                     & 96.12                     & 93.50                     & 61.02                     & 84.60                      \\
ResNet-50 \cite{resnet}   & 84.45                     & 91.98                     & 96.26                     & 89.09                     & 82.80                      & 74.72                     & 96.28                     & 93.88                     & 63.54                     & 85.80                      \\
ResNet-101 \cite{resnet}  & 85.58                     & 92.38                     & 96.53                     & 89.47                     & 84.88                     & 74.80                      & 97.39                     & 93.92                     & 63.76                     & 85.68                     \\
ResNet-152 \cite{resnet}  & 86.28                     & 93.10                      & 96.86                     & 89.88                     & 85.66                     & 76.44                     & 97.53                     & 94.42                     & 64.82                     & 86.32                     \\
DenseNet-121 \cite{densenet} & 84.99                     & 91.50                      & 97.02                     & 89.34                     & 82.75                     & 74.18                     & 96.45                     & 93.07                     & 63.26                     & 85.28                     \\
DenseNet-161 \cite{densenet} & 87.13 & 93.13 & 97.59 & 89.62 & 84.98 & 76.21 & 97.48 & 94.35 & 64.25 & 85.69 \\
DenseNet-169 \cite{densenet} & 85.84                     & 92.51                     & 97.32                     & 89.02                     & 84.26                     & 74.72                     & 96.77                     & 93.62                     & 64.10                      & 85.77                     \\
DenseNet-201 \cite{densenet} & 86.71                     & 93.14                     & 97.10                      & 89.44                     & 84.88                     & 76.04                     & 97.02                     & 94.03                     & 64.57                     & 85.67                     \\
MNet-A1  \cite{mnasnet}    & 71.35                     & 89.34                     & 95.39                     & 72.58                     & 72.04                     & 70.12                     & 92.59                     & 91.08                     & 56.56                     & 81.06                     \\
MobileNetV2 \cite{mobilenetv2} & 81.12                     & 88.64                     & 96.20                      & 86.44                     & 78.11                     & 71.72                     & 94.74                     & 91.28                     & 60.29                     & 82.80                      \\
Googlenet \cite{googlenet}   & 79.30                      & 90.85                     & 95.76                     & 87.76                     & 79.84                     & 72.53                     & 95.54                     & 91.38                     & 59.89                     & 82.58                     \\
InceptionV3 \cite{inceptionv3} & 81.76                     & 92.75                     & 95.73                     & 87.74                     & 81.49                     & 72.85                     & 96.18                     & 92.14                     & 59.98                     & 83.84                     \\
\bottomrule[1pt]
\end{tabular}%
}

\end{table*}

\begin{table*}[t]
\caption{The ground-truth results of the 12 self-supervised pre-trained models on 10 downstream tasks.}
\centering
\label{tab:gt_sl}
{
\begin{tabular}{ccccccccccc}
\toprule[1pt]
\multicolumn{1}{c}{Self-Supervised} & Food  & Caltech & Flowers & Cars & CIFAR100 & DTD   & CIFAR10 & Pets  & SUN397 & VOC2007 \\ \hline
BYOL \cite{byol}                                 & 85.44 & 91.90   & 96.80   & 89.83                     & 83.86    & 76.37 & 96.98   & 91.48 & 63.69  & 85.13   \\
Deepclusterv2 \cite{deepcluster}                 & 87.24 & 91.16   & 97.05   & 90.16                     & 84.84    & 77.31 & 97.17   & 90.89 & 66.54  & 85.38   \\
Infomin \cite{infomin}                           & 78.82 & 80.86   & 95.81   & 86.90                     & 70.89    & 73.74 & 96.72   & 90.92 & 57.67  & 81.41   \\
InsDis \cite{indis}                              & 76.47 & 77.21   & 93.63   & 80.21                     & 69.08    & 66.40 & 93.08   & 84.58 & 51.62  & 76.33   \\
MoCov1 \cite{moco-v1}                            & 77.21 & 79.68   & 94.32   & 82.19                     & 71.23    & 67.36 & 84.15   & 85.26 & 53.83  & 77.94   \\
MoCov2 \cite{moco-v2}                            & 77.15 & 82.76   & 95.12   & 85.55                     & 71.27    & 72.56 & 96.48   & 89.06 & 56.28  & 78.32   \\
PCLv1 \cite{pcl}                                 & 77.70 & 88.60   & 95.62   & 87.15                     & 79.44    & 73.28 & 86.42   & 88.93 & 58.36  & 91.91   \\
PCLv2 \cite{pcl}                                 & 80.29 & 87.52   & 95.87   & 85.56                     & 79.84    & 69.30 & 96.55   & 88.72 & 58.82  & 81.85   \\
Selav2 \cite{sela}                              & 86.37 & 90.53   & 96.22   & 89.85                     & 84.36    & 76.03 & 96.85   & 89.61 & 65.74  & 85.52   \\
SimCLRv1 \cite{simclr-v1}                        & 82.20 & 90.94   & 95.33   & 89.98                     & 84.49    & 73.97 & 97.09   & 88.53 & 63.46  & 83.29   \\
SimCLRv2 \cite{simclr-v2}                        & 82.23 & 88.58   & 95.39   & 88.82                     & 78.91    & 94.71 & 96.22   & 89.18 & 60.93  & 83.08   \\
SWAV \cite{swav}                                 & 87.22 & 89.49   & 97.11   & 89.81                     & 83.78    & 76.68 & 96.81   & 90.59 & 66.10  & 85.06  \\
\bottomrule[1pt]
\end{tabular}
}
\end{table*}

\section{Ground-truth Results}
We obtained the ground-truth results after model fine-tuning which employs a grid-search strategy, following the implementation of \cite{sfda,ped}. More details on this process are available in Sec. 4 of the main paper. In Tab. \ref{tab:gt_s} and \ref{tab:gt_sl}, we present the ground-truth results of 12 supervised pre-trained models and 12 self-supervised pre-trained models across 10 downstream tasks.

\end{document}